\documentclass{article}

\usepackage{booktabs} 
\usepackage[small]{caption}
\usepackage[hidelinks]{hyperref}
\usepackage{algpseudocode}

\usepackage{algorithmic}
\usepackage{graphicx}
\usepackage{subcaption}
\usepackage[accepted]{icml2024}
\usepackage{microtype}
\usepackage{graphicx}

\usepackage{refcount}
\usepackage{amsmath}
\usepackage{amssymb}
\usepackage{mathtools}
\usepackage{amsthm}

 \usepackage[normalem]{ulem}
\usepackage{soul}

\usepackage{times}
\usepackage{amsfonts}
\usepackage{multirow}
\usepackage{adjustbox}

\usepackage{url}
\usepackage{dsfont}
\usepackage[utf8]{inputenc}

\usepackage{amsmath}
\usepackage{booktabs}
\usepackage[most]{tcolorbox}
\hypersetup{
  colorlinks,
  citecolor=violet,
  linkcolor=red,
  urlcolor=blue}

\usepackage{lipsum}
\usepackage[capitalize,noabbrev]{cleveref}

\theoremstyle{plain}

\theoremstyle{definition}

\theoremstyle{remark}

\icmltitlerunning{A Novel Hyperdimensional Computing Framework for Online Time Series Forecasting on the Edge}

\begin{document}

\twocolumn[
\icmltitle{A Novel Hyperdimensional Computing Framework for Online Time Series Forecasting on the Edge}



\icmlsetsymbol{equal}{*}

\begin{icmlauthorlist}
\icmlauthor{Mohamed Mejri}{yyy}
\icmlauthor{Chandramouli Amarnath}{yyy}
\icmlauthor{Abhijit Chatterjee}{yyy}
\end{icmlauthorlist}

\icmlaffiliation{yyy}{School of Electrical And Computer Engineering, Georgia Institute of Technology, Atlanta, Georgia, US}

\icmlcorrespondingauthor{Mohamed Mejri}{mmejri3@gatech.edu}
\icmlcorrespondingauthor{Chandramouli Amarnath}{chandamarnath@gatech.edu}

\icmlkeywords{Machine Learning, ICML}

\vskip 0.3in
]



\printAffiliationsAndNotice{}  

\begin{abstract}
In recent years, both online and offline deep learning models have been developed for time series forecasting. However, offline deep forecasting models fail to adapt effectively to changes in time-series data, while online deep forecasting models are often expensive and have complex training procedures. In this paper, we reframe the online nonlinear time-series forecasting problem as one of linear \textit{hyperdimensional} time-series forecasting. Nonlinear low-dimensional time-series data is mapped to high-dimensional (hyperdimensional) spaces for linear hyperdimensional prediction, allowing fast, efficient and lightweight online time-series forecasting. Our framework, \textit{TSF-HD}, adapts to time-series distribution shifts using a novel co-training framework for its hyperdimensional mapping and its linear hyperdimensional predictor. TSF-HD is shown to outperform the state of the art, while having reduced inference latency, for both short-term and long-term time series forecasting. Our code is publicly available at: \href{https://github.com/tsfhd2024/tsf-hd.git}{https://github.com/tsfhd2024/tsf-hd.git}
\end{abstract}

\section{Introduction}
Time series forecasting methods have shown significant utility in fields ranging from smartgrids to traffic management \cite{HyndmanSurvey}, and this usefulness has driven increasing research into better forecasting models \cite{BhatnagarMerlion}. However, as noted in \cite{pham2022learning}, naive training of deep forecasting models or offline forecasting models may not generalize well to streams of time-series data, requiring the forecaster to be trained on-line. 
Recent works such as OneNet \cite{zhang2023onenet}, TCN \cite{woo2022cost} and FSNet \cite{pham2022learning} have thus focused on training online deep forecasting models that aim to use novel training paradigms to enable deep neural networks to rapidly assimilate information from a time-series data stream. 
 
However, these online deep forecasting models are expensive and difficult to deploy on edge platforms compared to linear time-series forecasters such as ARIMA \cite{box1968some}. Despite the success of deep forecasting models, it has also been shown in prior work \cite{zeng2023transformers} that certain linear models such as NLinear can outperform transformer methods and deep forecasting methods in long-term time-series forecasting. Our research is thus motivated by this tradeoff between performance and overhead between linear and nonlinear models in time-series forecasting. 


This work efficiently addresses that tradeoff by framing the forecasting problem as one of task-free online \textit{hyperdimensional} learning. Hyperdimensional computing is a learning paradigm that uses high-dimensional (hyperdimensional) mappings of nonlinear input data distributions to enable linear classification or regression \cite{kanerva2009hyperdimensional,hernandez2021reghd} in high dimensions. This leverages the fact that functions which are nonlinear in low dimensions can be approximated as linear in high dimensions while preserving distances \cite{mapping_cite_intro_rasanen}. Using this, we can run high-dimensional linear computations for nonlinear time-series forecasting, allowing rapid, low-overhead model updates from incoming samples in a data stream while maintaining state-of-the-art performance.

We thus propose \textit{TSF-HD}, an online hyperdimensional time-series forecasting framework leveraging the rapid training and inference capabilities of linear time-series forecasting models while allowing state-of-the-art prediction accuracy. This is accomplished using a novel online training method for the hyperdimensional computing system and implementing an innovative co-training method for the high-dimensional mapping (the \textit{encoder}) and the linear high-dimensional \textit{regressor}. This co-training of regressor and encoder allows us to maintain the accuracy of linear predictions made by the regressor as the nonlinear time series shifts. We emphasize that TSF-HD is a task-free, online learning model - it does not require explicit detection of task shifts (changes in time series distributions, as in \cite{pham2022learning}) or time series concept drift. TSF-HD instead learns online using current samples through trainable high-dimensional mappings of input data and high-dimensional linear regression. We further provide a method for autoregressive time-series hyperdimensional time-series forecasting, allowing greater accuracy over long prediction horizons or noisy time-series data by framing the problem as one of autoregressive online hyperdimensional time-series forecasting. For low-overhead, low-latency prediction we provide a sequence-to-sequence model of TSF-HD.

In summary, our work provides the following innovations: (1) \textit{First}, we provide a novel formulation of the time series forecasting problem as one of \textit{task free, online hyperdimensional learning}, taking advantage of the linearity of high dimensional mappings of nonlinear input data; (2) \textit{Second}, we provide a novel online co-training framework for the encoder and regressor to enable the linearity of the mapping to be maintained as the time series evolves, without requiring explicit knowledge of task shifts. This is enhanced by the provision of an autoregressive version of TSF-HD for greater precision on noisy data streams; (3) \textit{Lastly}, we conduct experiments against a variety of baselines to validate TSF-HD in terms of accuracy, latency and overhead.

The rest of the paper is organized as follows. Prior work is discussed in Section \ref{related_work}, followed by problem framing and methods in Section \ref{online}. In Section \ref{exp}, we present experimental results and finally conclude in Section \ref{conc}.
\section{Related work}\label{related_work}

\subsection{Time series forecasting (TSF)}
In recent years, advancements in data availability and computational power have led to the emergence of deep learning-based techniques in time series forecasting (TSF). Traditional methods such as Autoregressive Integrated Moving Average (ARIMA) \cite{box1968some}, Autoregressive Neural Networks (AR-Net) \cite{triebe2019ar}, the Holt-Winters seasonal method \cite{holt2004forecasting}, and Gradient Boosting Regression Trees (GBRT) \cite{friedman2001greedy} provide theoretical guarantees, but are typically applied to univariate time-series forecasting. Furthermore, these methods are often outperformed by deep forecasting models.

Recurrent Neural Network (RNN)-based approaches, such as those discussed by \cite{rangapuram2018deep}, are examples of deep forecasting models with internal memory that are able to make predictions based on past information. However, RNN-based models face challenges due to vanishing or exploding gradient problems and inefficient training procedures. Transformers, exemplified by the Informer model \cite{zhou2021informer}, have outperformed the RNN paradigm, leveraging the self-attention mechanism's capability to capture correlations in temporal sequences. SCINet \cite{liu2022scinet} is a convolutional deep forecaster that recursively downsamples and convolves data to extract complex temporal features for accurate time-series forecasting. Despite the success of transformers in time series forecasting, simpler linear models like NLinear \cite{zeng2023transformers} have outperformed transformer methods in some long-term time series forecasting applications in terms of accuracy.

However, none of the methods described above are adapted to online time series forecasting. Recent work in online time series forecasting includes an online Temporal Convolutional Network (TCN) with 10 hidden layers that successfully captures periodic patterns \cite{woo2022cost}. The Experience Replay method (ER) \cite{chaudhry2019tiny} enhances online TCN by adding episodic memory to mix old and new learning samples, improving generalization. DER++ \cite{buzzega2020dark} augments the standard ER with an $\ell_2$ knowledge distillation loss on the previous logits to align the network's logits, ensuring consistency with its past behavior. FSNet \cite{pham2022learning}, or Fast and Slow Learning Network, combines rapid adaptation to new data with memory recall of past events to enable adaptation to task shifts (concept drift) in the time series. OneNet \cite{zhang2023onenet} extends the FSNet implementation 
by integrating and updating two models in real-time: one concentrates on modeling dependencies over time, and the other focuses on dependencies across variables.
We note that these online deep forecasters are expensive and cumbersome to train, and may not be applicable to low-latency forecasting in a shifting nonlinear time series stream.

Using high-dimensional (hyperdimensional) mappings (encodings) to forecast a nonlinear data stream using linear hyperdimensional regression has not been explored in prior work, primarily due to the lack of a co-trainable encoder and regressor in state of the art hyperdimensional regressors like RegHD \cite{hernandez2021reghd,chen2022full}. This deficiency prevents them from adapting to time-series task shifts and learning online in an effective manner.

\subsection{Online learning}
Online learning \cite{NIPS2017_f8752278} aims to learn several tasks sequentially with limited access to past experiences. A good learner is one that achieves the best trade-off between adaptation to new tasks and maintenance of past knowledge from previous tasks, a tradeoff known as the stability-plasticity dilemma \cite{grossberg1982does}. 
One popular framework is the complementary learning system (CLS) \cite{mcclelland1995there,kumaran2016learning}. Continual deep learning methods using the CLS framework augment slow deep learning algorithms with the ability to learn quickly from a data stream, either through experience replay \cite{lin1992self,riemer2018learning,rolnick2019experience,aljundi2019task,buzzega2020dark} or via fast and slow learning components \cite{pham2020contextual,arani2022learning,pham2022learning}. 

\begin{figure}[htb]
\centering
    \begin{subfigure}[b]{0.4\textwidth}
        \centering
        \includegraphics[width=\textwidth]{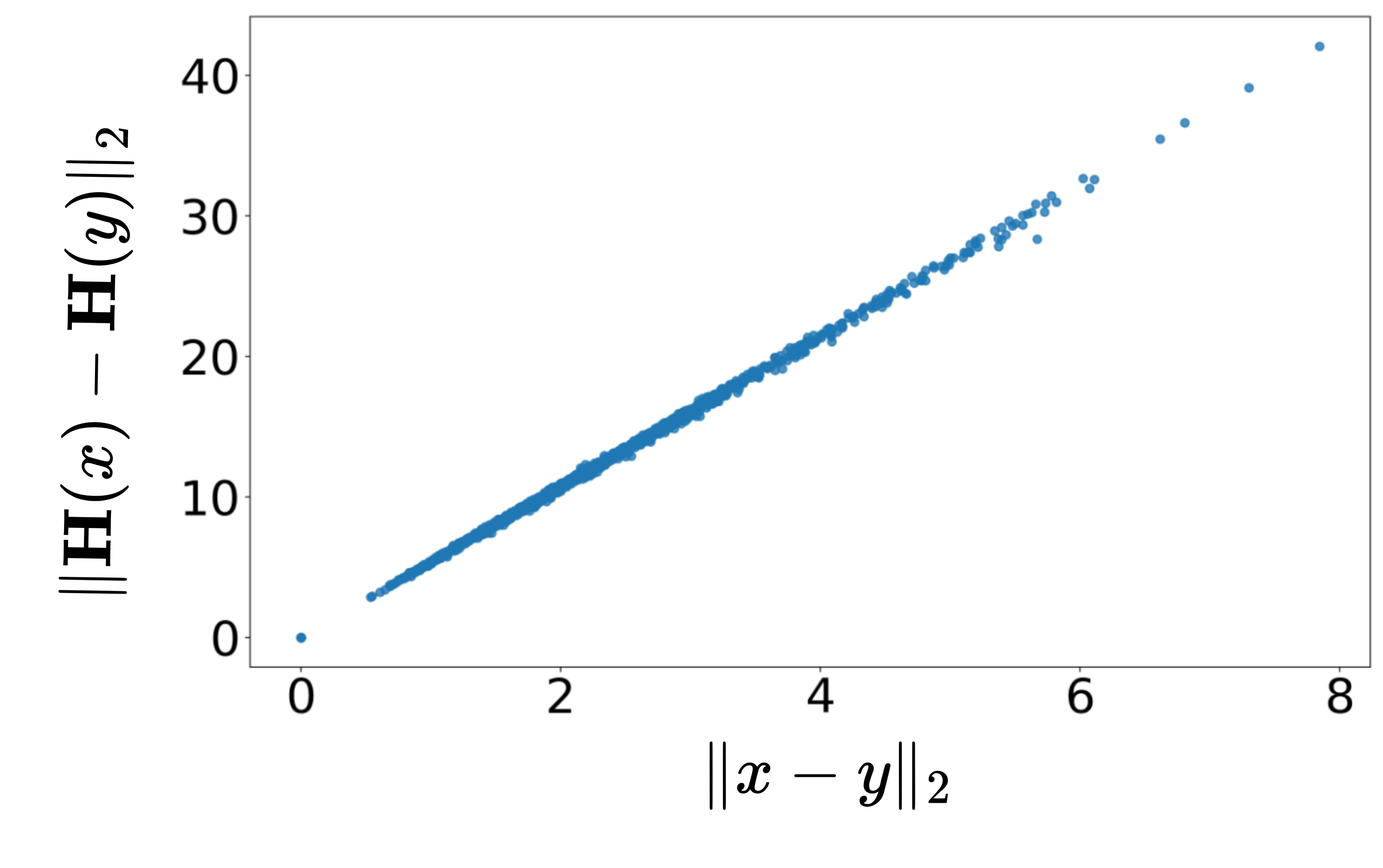}
        \caption{Distance preservation across hyperdimensional mappings of input sequences $\mathbf{H}(x)$.}
        \label{distance_1}
    \end{subfigure}

    \begin{subfigure}[b]{0.4\textwidth}
        \centering
        \includegraphics[width=\textwidth]{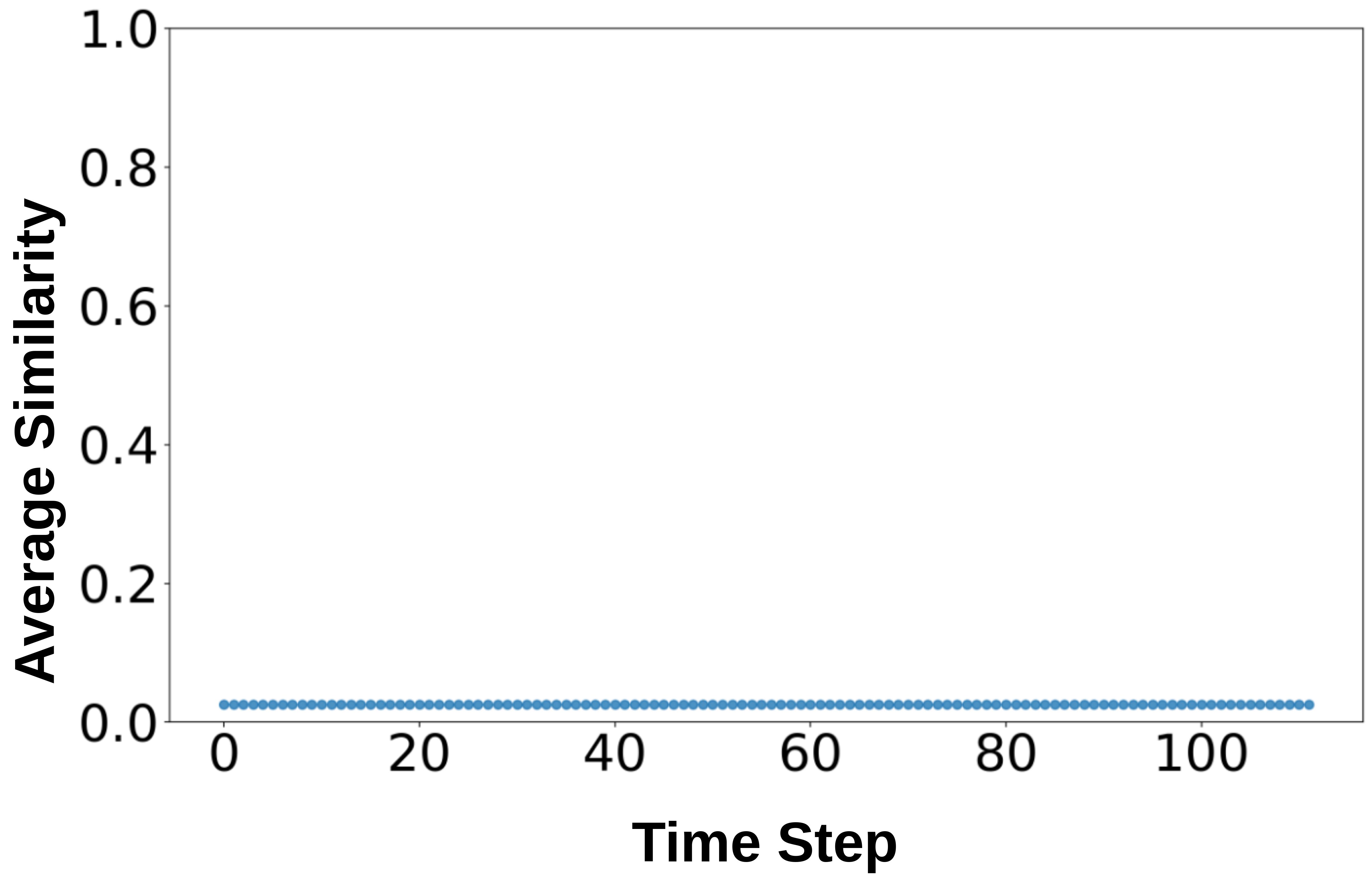}
        \caption{The mean cosine similarity between rows of the encoder matrix $W_e$ used to map low-dimensional input sequences to the hyperdimensional space is seen to be very low as the time series evolves ($t$ increases), preserving orthogonality while updating $W_e$ online.}
        \label{distance_2}
    \end{subfigure}
    \caption{Distance preservation (linear relationship between distances) and orthogonality of the different components of the trainable encoder matrix $W_e$ across hyperdimensional mappings of input sequences $\mathbf{H}(x)=x.W_e$ (ETTm1 dataset), showing preservation of the properties detailed in \cite{mapping_cite_intro_rasanen}.}
    \vspace{-0.5cm}
\end{figure}

\section{Online Time Series Forecasting}\label{online}
\subsection{Problem Framing and Preliminaries}\label{prelims}

In this paper, we focus on the \textit{online} multivariate time series forecasting (TSF) problem. Given a long sequence $\mathbf{X}^* \in \mathbb{R}^{N \times L}$ where $L$ is the length of the sequence $\mathbf{X}^*$ and $N$ is the number of variables in the time series $\mathbf{X}^*$ (the dimension of the input $x\in\mathbf{X}^*$) and given a look back window of fixed length $T$, at timestamp $\textit{t}$, the time series forecasting task is to predict $X_{t+1:t+\tau}=[x_{t+1},...,x_{t+\tau} ]$ based on the past T steps $X_{t-T+1:t}=[x_{t-T+1},...,x_{t}]\in\mathbb{R}^{N\times T}$. Here $\tau$ is the referred to as the \textit{forecast horizon}. Each data sample $x \in \mathbb{R}^{N}$. Online learning assumes that data arrives in a stream, drawn from a shifting distribution.
Our approach leverages the ability of hyperdimensional spaces to represent nonlinear input data in low dimensions as linear in high dimensions \cite{mapping_cite_intro_rasanen}. The learner is not aware when the data changes in trend (i.e, moves from one task to another) making the online time series forecasting problem one of task-free online learning. We thus use hyperdimensional regression with a trainable hyperdimensional mapping (the \textit{encoder}) and a novel co-training framework for the encoder and the linear hyperdimensional regressor.

Hyperdimensional (HD) computing represents data using extremely high dimensional spaces \cite{kanerva2009hyperdimensional} referred to as `hyperspaces'. A vector in this space is referred to as a `hypervector'. Multivariate hyperdimensional regression in TSF-HD involves three primary phases:

    \noindent\textbf{Hypervector Encoding:} In this phase, an input data sequence $X\in\mathbb{R}^{N\times T}$ is transformed from the feature space $\mathcal{X}$ to a hyperspace $\mathcal{H}$ ($dim(\mathcal{H})=D \gg T$) using a function $\mathbf{H}: \mathcal{X} \rightarrow \mathcal{H}$. The encoding can be accomplished using various methods such as N-gram based encoders \cite{imani2018hierarchical}, or linear projection \cite{dutta2022hdnn}. Our work uses a \textit{trainable} encoder consisting of a linear mapping followed by a ReLU function, yielding 
    \begin{equation}\label{encoding}
        \mathbf{H}(X) = \mathds{1}_{X>0}[X.W_e + b_e]
    \end{equation}
    where $W_e\in\mathbb{R}^{T\times D}$ is a matrix whose rows consist of hypervectors mapping the input $X$ to different components of the HD space, $\mathds{1}_{X>0}(.)$ denotes the ReLU function and $b_e \in \mathbb{R}^{D}$ is a trainable bias added to each row of $X.W_e$.
    
    As per \cite{mapping_cite_intro_rasanen}, based on the Johnson-Lindenstrauss lemma, the distances between $x$ and $y$ ($x,y\in\mathcal{X}$) are preserved under the encoder mapping $H(.)$ within a scaling factor for a \textit{non-trainable}, randomized projection matrix mapping $\mathcal{X}$ to $\mathcal{H}$, allowing an HD system to take advantage of the linearity of high dimensional mappings of lower dimensional nonlinear input data. To enable forecasting on a shifting time series, our \textit{trainable} encoder system evolves with the time series sequence while still preserving distances, as seen in Figure \ref{distance_1}, where the plot between the distances $\lVert x - y\rVert_2$ and $\lVert \mathbf{H}(x)-\mathbf{H}(x)\rVert_2$ is linear, where $\lVert.\rVert_2$ denotes the L2 norm. Similarly, as per \cite{mapping_cite_intro_rasanen}, for a \textit{non-trainable} randomized projection matrix mapping $\mathcal{X}$ to $\mathcal{H}$, the hypervectors that make up the matrix are ideally orthogonal to one another. We see that this approximately holds through the time series in Figure \ref{distance_2} for our trainable matrix $W_e$, with the average cosine similarity between its rows remaining near-zero. Further validation of distance preservation and orthogonality of $W_e$ can be found in Appendix \ref{appendix_framing}. Hyperdimensional computing using our novel trainable encoder and regressor formulation thus allows us to take advantage of the desirable properties of hyperdimensional encoding detailed in \cite{mapping_cite_intro_rasanen} while continually updating our model as the data stream shifts.
    
    \noindent\textbf{HD System Training:} For our linear HD regressor, the goal is to find an approximation $\tilde{X}_{t+1:t+\tau}$, given an \textit{encoded} input sequence taken from the lookback window, $\mathbf{H}(X_{t-T+1:t})$ to minimize a loss function $\mathcal{L}(.)$ calculated from the regression error across the prediction horizon $(\tilde{X}_{t+1:t+\tau}-X_{t+1:t+\tau})$. This involves a trainable regressor hypervector or matrix, denoted as $W_r$, and a trainable regressor bias $b_r$. $W_r$ and $b_r$ are updated in conjunction with $W_e$ and $b_e$ to minimize $\mathcal{L}(\tilde{X}_{t+1:t+\tau}-X_{t+1:t+\tau})$, using online gradient descent (OGD) with the AdamW optimizer.

    Our input data is not normalized or standardized, simulating a real-time online learning environment, and data point values may vary significantly. An L2-norm-based loss function could lead to rapid divergence of the loss due to such variations. While an L1 norm is more suitable, its non-differentiability around 0 presents a challenge. Therefore, we opt for the \textit{Huber loss} ($\mathcal{L}_{H}$), as detailed for a single prediction step ($\tau=1$) in Eq. \ref{HL}. We denote $\Delta x_{t+1} = x_{t+1} - \tilde{x}_{t+1}$, yielding:
\begin{equation}
\begin{aligned}
    \mathcal{L}_H(x_{t+1}, \tilde{x}_{t+1}) = 
    \begin{cases}
    \frac{1}{2} \lVert \Delta x_{t+1}\rVert_2, & \text{if } \lvert \Delta x_{t+1} \rvert \leq 1 \\
    \lVert \Delta x_{t+1} \rVert_1 - \frac{1}{2}, & \text{otherwise}
    \end{cases}
\end{aligned}
\label{HL}
\end{equation}
    The total loss is thus
    \begin{equation}
        \mathcal{L}(x_{t+1},\tilde{x}_{t+1}) = \mathcal{L}_H(x_{t+1},\tilde{x}_{t+1}) + \mathcal{R}(W_e,W_r,b_e,b_r) 
    \end{equation}
    where $\mathcal{R}(.)$ is an L2 norm regularization function.
    
    \noindent\textbf{HD Inference} The prediction $\tilde{X}_{t+1:t+\tau}$ is generated after mapping the input samples $X_{t-T+1:t}$ to the hyperspace $\mathcal{H}$ as in Equation \ref{encoding}. A single-step prediction $\tilde{x}_{t+1}$ is then computed as:
    \begin{equation}\label{regression}
    \begin{split}
    \tilde{x}_{t+1} = \mathbf{R}(\mathbf{H}(X_{t-T+1:t})) = [(\left(\mathbf{H}(X_{t-T+1:t})\right)_1.W_r,... \\ ...,\left(\mathbf{H}(X_{t-T+1:t})\right)_N.W_r ]+b_r
    \end{split}
    \end{equation}
    where $\mathbf{H}(.)$ is the encoding function of Equation \ref{encoding} and $\mathbf{R}(.)$ is the regression function that uses the regression matrix $W_r$ and bias $b_r$. $\left(\mathbf{H}(X_{t-T+1:t})\right)_i$ denotes the $i$th row of an $N\times D$-dimensional encoded input sequence. The regressor function for the $i$th element of the predicted vector ($1\leq i\leq N$) is thus $(\tilde{x}_{t+1})_i=\mathbf{R}_i(\mathbf{H}(X_{t-T+1:t})) = \left(\mathbf{H}(X_{t-T+1:t})\right)_i.W_r+b_r$
We provide two distinct TSF-HD frameworks: the autoregressive AR-HDC and the sequence-to-sequence Seq2Seq-HDC. AR-HDC is more accurate than Seq2Seq-HDC, but incurs higher overhead and is slower for long-term forecasting than Seq2Seq-HDC. 

\begin{figure}
    \centering
    \includegraphics[width=\columnwidth]{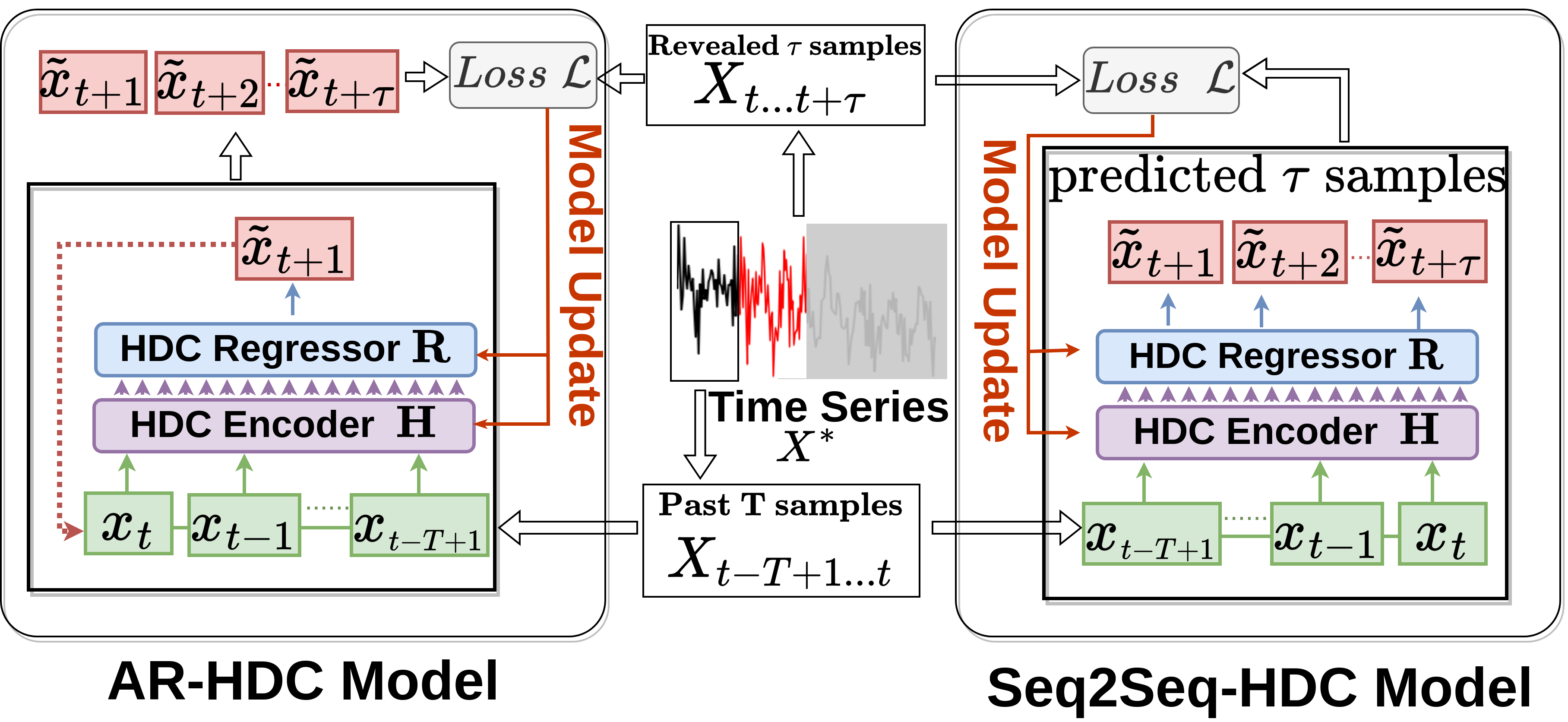}
    \caption{Autoregressive (AR-HDC) \& Sequence-to-Sequence (Seq2Seq) frameworks for TSF-HD. Our novel co-training system jointly updates the encoder $\mathbf{H}(.)$ and regressor $\mathbf{R}(.)$ online.}
    \label{fig:ar-seq-overview}
    \vspace{-0.5cm}
\end{figure}

\subsection{TSF-HDC Frameworks}
\subsubsection{Overview}

Figure \ref{fig:ar-seq-overview} presents an overview of both the autoregressive version of our framework, called AR-HDC, and the sequence-to-sequence version, called Seq2Seq-HDC. Both models use $T$ multivariate past samples $X_{t-T+1...t} \in \mathbb{R}^{N\times T}$, to predict the next $\tau$ samples, $\tilde{x}_{t+1}$ to $\tilde{x}_{t+\tau}$.

The autoregressive AR-HDC framework (Figure \ref{fig:ar-seq-overview}, left) predicts one time step ahead at a time. This prediction is then used to make further predictions, as seen in Figure \ref{fig:ar-seq-overview} when $\tilde{x}_{t+1}$ is fed back - the system thus predicts $\tilde{x}_{t+1}$ using $X_{t-T+1:t}$, then predicts $\tilde{x}_{t+2}$ using $[X_{t-T+2:t},\tilde{x}_{t+1}]$, until it predicts $\tilde{x}_{t+\tau}$ using $[X_{t-T+\tau+1:t},\tilde{X}_{t+1:t+\tau-1}]$. Once the prediction is complete, the system uses the true values of the $\tau$ samples ($x_{t+1}$ to $x_{t+\tau}$) to update the model - model updates are thus done periodically once new information is known.
AR-HDC then attempts to minimize the loss $\mathcal{L}$ by updating the weights of its components (the regressor and encoder), as in Figure \ref{fig:ar-seq-overview}. Unlike prior work, the regressor and encoder are updated \textit{jointly} online to minimize $\mathcal{L}$.

In contrast, the Seq2Seq-HDC system (Figure \ref{fig:ar-seq-overview}, right) predicts the future sample values (from $\tilde{x}_{t+1}$ to $\tilde{x}_{t+\tau}$) in one shot using the values of $X_{t-T+1:t}$, as seen in Fig. \ref{fig:ar-seq-overview}. This is faster than the iterative prediction of AR-HDC, but may not be as accurate. As seen in Figure \ref{fig:ar-seq-overview}, the true values of the time points in the prediction horizon (ranging from $t+1$ to $t+\tau$) are then used to calculate the prediction loss across the sequence and update the model, again updating regressor and encoder jointly (co-training) to minimize $\mathcal{L}$.

\subsubsection{Autoregressive HDC (AR-HDC)}
AR-HDC predicts one element ahead at a time, modeling the time series as an autoregressive process. The classical autoregressive predictor yields the predicted multivariate sample $\tilde{x}_{t+1} \in \mathbb{R}^{N}$ using a weighted average of past $T$ (i.e, the look back window value) samples $X_{t-T+1:t}$, yielding 
    $\tilde{x}_{t+1}=\sum_{j=0}^{T} w_{j}.x_{t-j+T}$.

For AR-HDC, the weighted average of $x_{t-j+T}$ of the previous equation is replaced with an hyperdimensional encoder and regressor (rewriting Equation \ref{regression}):
\begin{equation}
    \left(\tilde{x}_{t+1}\right)_i=<W_r,\left(\mathbf{H}(X_{t-T+1:t})\right)_i>+b_r
    \label{AR-HD-eq}
\end{equation}
where $<.>$ denotes the inner product operation, $W_r \in \mathbb{R}^{D}$ denotes the trainable regressor hypervector (described in Section \ref{prelims}), $\left(\tilde{x}_{t+1}\right)_i$ is the $i$th term of the predicted vector ($1\leq i\leq N)$, $b_r$ is the regressor bias term and $\mathbf{H}(X_{t-T+1:t})_i$ denotes $i$th row of the encoded input sequence $X_{t-T+1:t}$. AR-HDC then adds $\tilde{x}_{t+1}$ to its lookback window and removes the last term from the lookback window to predict $x_{t+2}$ similarly, using the sample vector $[X_{t-T+2:t},\tilde{x}_{t+1}]$. This is described in Algorithm \ref{AR-HDC-end}.

\begin{algorithm}
\caption{Autoregressive HDC (AR-HDC)}
\label{AR-HDC-end}
\begin{algorithmic}[1]
\State $X \gets X_{t-T+1:t}=\left[x_{t-T+1}, \ldots, x_{t-1}, x_{t}\right]$
\For{$i = 1$ \textbf{to} $\tau$}
        \State $\tilde{x}_{t+i} \gets \mathbf{R}(\mathbf{H}(X))$
        \State $X \gets  (X \setminus \{x_{t-T+i}\}) \cup \{\tilde{x}_{t+i}\}$
    \EndFor
\For{$i = 1$ \textbf{to} $\tau$}
 \State \Call{OGD}{$\mathbf{H}, \mathbf{R}, \mathcal{L}(\tilde{x}_{t+i}, x_{t+i})$} \Comment{Training Step}
    \EndFor

\end{algorithmic}
\end{algorithm}

The procedure for AR-HDC consists of two phases: a prediction phase for $\tau$ future samples (from line 2 to line 5 of Algorithm \ref{AR-HDC-end}) and an online learning phase (from line 6 to line 8) to update the encoder and regressor once the true values of $X_{t+1:t+\tau}$ are known. At line 3, AR-HDC employs the encoding $\mathbf{H}$, followed by the regression function $\mathbf{R}$, to predict the $i^{th}$ future sample based on the past sequence $X$, which is initialized in line 1 as the set of samples $X=X_{t-T+1:t}$. In line 4, the first element $x_{t-T+i}$ (initially $x_{t-T+1}$) of the sequence $X$ is removed, and the predicted sample $\tilde{x}_{t+i}$ is appended. This updated sequence is then fed forward to the encoder and regressor in line 2. This is repeated until $x_{t+\tau}$ is predicted from $X=[X_{t-T+\tau+1:t},\tilde{X}_{t+1:t+\tau-1}]$. Following the prediction phase, the system parameters (encoder weights and regression hypervectors) are updated using online gradient descent as in Section \ref{prelims}. The overhead of AR-HDC thus scales linearly with $\tau$. For large $\tau$, a sequence-to-sequence predictor may thus be more practical.
\subsubsection{Sequence-To-Sequence HDC (Seq2Seq-HDC)}
The Seq2Seq-HDC algorithm is an efficient and straightforward algorithm for time series forecasting, where the past $T$ steps of the sequence are encoded into a $D\gg T$-dimensional hyperspace and a linear HDC regressor consisting of a $D\times \tau$-dimensional regression matrix and trainable bias $b_r$ is used. The HDC encoder $\mathbf{H}$ is identical to that of AR-HDC. The system begins by generating hypervector encodings $h=\mathbf{H}(X_{t-T+1:t})$, which are then used by the regressor $\mathbf{R}(.)$:
\begin{align}
    \tilde{X}_{t+1:t+\tau} = \mathbf{R}(h) = h.W_{r} + b_{r}
\end{align}
Here, $X_{t-T+1..t} \in \mathbb{R}^{N\times T}$ is the input sequence of past elements. It is encoded into an hyperdimensional structure $h \in \mathbb{R}^{N\times D}$ using $W_{e}$ and $b_{e}$ as in Equation \ref{encoding}. $W_{r} \in \mathbb{R}^{D\times\tau}$ is the HDC regression matrix for the output sequence $\tilde{X}_{t+1:t+\tau}\in\mathbb{R}^{N\times\tau}$ and $b_{r} \in \mathbb{R}^{\tau}$ is the regression bias. This process is further described in Algorithm \ref{algo_seq2seq}.

\begin{algorithm}
\caption{Seq2Seq-HDC Algorithm}
\begin{algorithmic}[1]
\State \textbf{Initialization:}
\State $X_{t-T+1..t} \gets \left[x_{t-T+1}, \ldots, x_{t-1}, x_{t}\right]$
\State \textbf{Forward Pass:}
\State $h \gets \mathbf{H}(X)$ 
\State $\tilde{X}_{t..t+\tau} \gets \mathbf{R}(h)$ 

\State \textbf{Online Learning Phase:}
\State $\Call{OGD}{(\mathbf{H}, \mathbf{R}, \mathcal{L}(X_{t..t+\tau}, \tilde{X}_{t..t+\tau}))}$ \Comment{Training Step}
\end{algorithmic}\label{algo_seq2seq}
\end{algorithm}
In line 4 the input, consisting of $T$ past elements $X_{t-T+1:t}$ are encoded into high dimensional vectors $h \in \mathbb{R}^{N \times D}$ using the encoder $\mathbf{H}$. In line 5, the hypervectors $h$ are forwarded to the regressor to generate the predicted $\tau$ future points $\tilde{X}_{t+1:t+\tau}$ using $\mathbf{R}(.)$. In the line 7, the actual values $X_{t+1:t+\tau}$ are used to update the system weights (encoder and all regressor matrices and biases) in order to minimize the loss $\mathcal{L}(X_{t+1:t+\tau},\tilde{X}_{t+1:t+\tau})$ via online gradient descent. 

\begin{table*}[ht]
\begin{center}
\begin{adjustbox}{center, width=0.75\textwidth}
\begin{tabular}{cc|cccccccccccccc|}
\cline{3-16}
 &
   &
  \multicolumn{2}{c}{{\color[HTML]{5B277D} \textbf{Seq2Seq-HDC}}} &
  \multicolumn{2}{c}{{\color[HTML]{5B277D} \textbf{AR-HDC}}} &
  \multicolumn{2}{c}{{\color[HTML]{ACB20C} \textbf{Fsnet}}} &
  \multicolumn{2}{c}{{\color[HTML]{ACB20C} \textbf{ER}}} &
  \multicolumn{2}{c}{{\color[HTML]{ACB20C} \textbf{DR++}}} &
  \multicolumn{2}{c}{{\color[HTML]{ACB20C} \textbf{OnlineTCN}}} &
  \multicolumn{2}{c|}{{\color[HTML]{2A6099} \textbf{Informer}}} \\ \cline{2-16} 
\multicolumn{1}{c|}{} &
  \textit{$\tau$} &
  \textit{\textbf{RSE}} &
  \textit{\textbf{CORR}} &
  \textit{\textbf{RSE}} &
  \textit{\textbf{CORR}} &
  \textit{\textbf{RSE}} &
  \textbf{CORR} &
  \textit{\textbf{RSE}} &
  \textit{\textbf{CORR}} &
  \textit{\textbf{RSE}} &
  \textit{\textbf{CORR}} &
  \textit{\textbf{RSE}} &
  \textit{\textbf{CORR}} &
  \textit{\textbf{RSE}} &
  \textit{\textbf{CORR}} \\ \hline
\multicolumn{1}{|c|}{} &
  3 &
  \textbf{0.032} &
  \textbf{0.999} &
  {\color[HTML]{3465A4} {\underline{0.033}}} &
  \textbf{0.999} &
  0.339 &
  0.916 &
  0.14 &
  0.99 &
  0.109 &
  {\color[HTML]{3465A4} {\underline{0.995}}} &
  0.194 &
  0.973 &
  0.682 &
  0.813 \\
\multicolumn{1}{|c|}{} &
  6 &
  {\color[HTML]{3465A4} {\underline{0.043}}} &
  \textbf{0.999} &
  \textbf{0.034} &
  \textbf{0.999} &
  0.516 &
  0.835 &
  0.148 &
 {\color[HTML]{3465A4} {\underline{0.987}}}&
  0.159 &
  {\color[HTML]{3465A4} {\underline{0.987}}} &
  0.175 &
  0.98 &
  0.704 &
  0.758 \\
\multicolumn{1}{|c|}{\multirow{-3}{*}{\textbf{Exchange}}} &
  12 &
  {\color[HTML]{3465A4} {\underline{0.053}}} &
  {\color[HTML]{3465A4} {\underline{0.998}}} &
  \textbf{0.035} &
  \textbf{0.999} &
  0.497 &
  0.872 &
  0.171 &
  0.982 &
  0.188 &
  0.979 &
  0.274 &
  0.95 &
  0.758 &
  0.754 \\ \hline
\multicolumn{1}{|c|}{} &
  3 &
  0.314 &
  0.952 &
  0.288 &
  0.966 &
  0.231 &
  0.966 &
  {\color[HTML]{3465A4} {\underline{0.194}}} &
  {\color[HTML]{3465A4} {\underline{0.978}}} &
  \textbf{0.181} &
  \textbf{0.981} &
  0.218 &
  0.97 &
  0.998 &
  0.153 \\
\multicolumn{1}{|c|}{} &
  6 &
  \textbf{0.153} &
  \textbf{0.988} &
  0.305 &
  0.976 &
  0.223 &
  0.971 &
  0.174 &
  0.977 &
  {\color[HTML]{3465A4} {\underline{0.168}}} &
  {\color[HTML]{3465A4} {\underline{0.982}}} &
  0.186 &
  0.977 &
  0.998 &
  0.137 \\
\multicolumn{1}{|c|}{\multirow{-3}{*}{\textbf{ECL}}} &
  12 &
  \textbf{0.106} &
  \textbf{0.995} &
  {\color[HTML]{3465A4} {\underline{0.121}}} &
  {\color[HTML]{3465A4} {\underline{0.994}}} &
  0.218 &
  0.979 &
  0.137 &
  0.992 &
  0.136 &
  0.992 &
  0.146 &
  0.991 &
  0.999 &
  0.046 \\ \hline
\multicolumn{1}{|c|}{} &
  3 &
  0.142 &
  0.989 &
  \textbf{0.128} &
  \textbf{0.994} &
  0.155 &
  {\color[HTML]{3465A4} {\underline{0.993}}} &
  0.141 &
  {\color[HTML]{3465A4} {\underline{0.993}}} &
  0.136 &
  \textbf{0.994} &
  0.164 &
  0.992 &
  0.751 &
  0.838 \\
\multicolumn{1}{|c|}{} &
  6 &
  0.162 &
  0.987 &
  0.177 &
  \textbf{0.991} &
  0.191 &
  {\color[HTML]{3465A4} {\underline{0.989}}} &
  {\color[HTML]{3465A4} {\underline{0.158}}} &
  \textbf{0.991} &
  \textbf{0.154} &
  \textbf{0.991} &
  0.191 &
  0.986 &
  0.757 &
  0.835 \\
\multicolumn{1}{|c|}{\multirow{-3}{*}{\textbf{ETTh2}}} &
  12 &
  0.178 &
  0.984 &
  \textbf{0.172} &
  \textbf{0.988} &
  0.229 &
  0.983 &
  0.197 &
  0.986 &
  0.188 &
  0.987 &
  0.226 &
  0.983 &
  0.753 &
  0.834 \\ \hline
\multicolumn{1}{|c|}{} &
  3 &
  {\color[HTML]{3465A4} {\underline{0.389}}} &
  \textbf{0.928} &
  \textbf{0.349} &
  {\color[HTML]{3465A4} {\underline{0.92}}} &
  0.509 &
  0.907 &
  0.652 &
  0.821 &
  0.513 &
  0.877 &
  0.583 &
  0.871 &
  0.947 &
  0.658 \\
\multicolumn{1}{|c|}{} &
  6 &
  {\color[HTML]{3465A4} {\underline{0.482}}} &
  0.894 &
  \textbf{0.443} &
  0.875 &
  0.609 &
  0.882 &
  0.489 &
  {\color[HTML]{3465A4} {\underline{0.906}}} &
  0.459 &
  \textbf{0.917} &
  0.538 &
  0.899 &
  0.99 &
  0.652 \\
\multicolumn{1}{|c|}{\multirow{-3}{*}{\textbf{ETTh1}}} &
  12 &
  \textbf{0.371} &
  \textbf{0.938} &
  {\color[HTML]{3465A4} {\underline{0.448}}} &
  {\color[HTML]{3465A4} {\underline{0.902}}} &
  0.727 &
  0.843 &
  0.533 &
  0.896 &
  0.521 &
  0.899 &
  0.597 &
  0.878 &
  1.005 &
  0.646 \\ \hline
\multicolumn{1}{|c|}{} &
  3 &
  {\color[HTML]{3465A4} {\underline{0.111}}} &
  \textbf{0.994} &
  \textbf{0.11} &
  \textbf{0.994} &
  0.198 &
  {\color[HTML]{3465A4} {\underline{0.987}}} &
  0.351 &
  0.914 &
  \textit{0.563} &
  \textit{0.821} &
  0.621 &
  0.892 &
  0.501 &
  0.887 \\
\multicolumn{1}{|c|}{} &
  6 &
  \textbf{0.135} &
  \textbf{0.991} &
  {\color[HTML]{3465A4} {\underline{0.148}}} &
  {\color[HTML]{3465A4} {\underline{0.99}}} &
  0.223 &
  0.985 &
  0.335 &
  0.954 &
  0.299 &
  0.952 &
  0.353 &
  0.911 &
  0.506 &
  0.884 \\
\multicolumn{1}{|c|}{\multirow{-3}{*}{\textbf{ETTm1}}} &
  12 &
  \textbf{0.181} &
  \textbf{0.984} &
  {\color[HTML]{3465A4} {\underline{0.209}}} &
  {\color[HTML]{3465A4} {\underline{0.982}}} &
  0.249 &
  0.981 &
  0.276 &
  0.965 &
  0.286 &
  0.956 &
  0.332 &
  0.941 &
  0.509 &
  0.883 \\ \hline
\multicolumn{1}{|c|}{} &
  3 &
  {\color[HTML]{3465A4} {\underline{0.104}}} &
  {\color[HTML]{3465A4} {\underline{0.994}}} &
  \textbf{0.086} &
  \textbf{0.998} &
  0.143 &
  0.995 &
  0.149 &
  0.983 &
  \textit{0.135} &
  \textit{0.986} &
  0.286 &
  0.951 &
  0.911 &
  0.795 \\
\multicolumn{1}{|c|}{} &
  6 &
  {\color[HTML]{3465A4} {\underline{0.136}}} &
  {\color[HTML]{3465A4} {\underline{0.991}}} &
  \textbf{0.114} &
  \textbf{0.996} &
  0.163 &
  0.994 &
  0.135 &
  0.992 &
  0.141 &
  0.991 &
  0.177 &
  0.991 &
  0.766 &
  0.904 \\
\multicolumn{1}{|c|}{\multirow{-3}{*}{\textbf{ETTm2}}} &
  12 &
  \textbf{0.161} &
  0.987 &
  {\color[HTML]{3465A4} {\underline{0.171}}} &
  \textbf{0.993} &
  0.155 &
  {\color[HTML]{3465A4} {\underline{0.992}}} &
  0.162 &
  {\color[HTML]{3465A4} {\underline{0.992}}} &
  0.148 &
  {\color[HTML]{3465A4} {\underline{0.992}}} &
  0.174 &
  0.991 &
  0.781 &
  0.901 \\ \hline
\multicolumn{1}{|c|}{} &
  3 &
  {\color[HTML]{000000} 0.671} &
  0.784 &
  {\color[HTML]{3465A4} {\underline{0.642}}} &
  0.797 &
  0.719 &
  0.811 &
  0.694 &
  {\color[HTML]{000000} 0.817} &
  0.682 &
  {\color[HTML]{3465A4} {\underline{0.819}}} &
  0.719 &
  0.809 &
  \textbf{0.58} &
  \textbf{0.831} \\
\multicolumn{1}{|c|}{} &
  6 &
  0.712 &
  0.773 &
  0.674 &
  0.785 &
  0.757 &
  0.807 &
  {\color[HTML]{000000} 0.708} &
  {\color[HTML]{3465A4} {\underline{0.819}}} &
  {\color[HTML]{3465A4} {\underline{0.697}}} &
  \textbf{0.822} &
  0.732 &
  0.812 &
  \textbf{0.612} &
  {\color[HTML]{000000} 0.811} \\
\multicolumn{1}{|c|}{\multirow{-3}{*}{\textbf{WTH}}} &
  12 &
  {\color[HTML]{3465A4} {\underline{0.665}}} &
  0.799 &
  \textbf{0.651} &
  0.802 &
  0.784 &
  0.8 &
  0.732 &
  {\color[HTML]{3465A4} {\underline{0.815}}} &
  {\color[HTML]{111111} 0.722} &
  \textbf{0.818} &
  0.759 &
  0.808 &
  0.655 &
  0.772 \\ \hline
\multicolumn{1}{|c|}{} &
  3 &
  0.203 &
  0.979 &
  \textbf{0.135} &
  \textbf{0.998} &
  0.143 &
  0.996 &
  0.151 &
  0.996 &
  {\color[HTML]{3465A4} {\underline{0.148}}} &
  {\color[HTML]{3465A4} {\underline{0.997}}} &
  0.158 &
  0.996 &
  1.187 &
  0.348 \\
\multicolumn{1}{|c|}{} &
  6 &
  0.247 &
  0.969 &
  0.202 &
  \textbf{0.995} &
  0.202 &
  0.993 &
  {\color[HTML]{3465A4} {\underline{0.199}}} &
  {\color[HTML]{3465A4} {\underline{0.994}}} &
  \textbf{0.189} &
  {\color[HTML]{3465A4} {\underline{0.994}}} &
  {\color[HTML]{3465A4} {\underline{0.199}}} &
  {\color[HTML]{3465A4} {\underline{0.994}}} &
  1.168 &
  0.373 \\
\multicolumn{1}{|c|}{\multirow{-3}{*}{\textbf{ILI}}} &
  12 &
  0.296 &
  0.956 &
  0.297 &
  \textbf{0.989} &
  \textbf{0.248} &
  \textbf{0.989} &
  {\color[HTML]{3465A4} {\underline{0.268}}} &
  {\color[HTML]{3465A4} {\underline{0.987}}} &
  0.328 &
  0.985 &
  0.279 &
  {\color[HTML]{3465A4} {\underline{0.987}}} &
  1.066 &
  0.337 \\ \hline
\end{tabular}
\end{adjustbox}
\end{center}
\caption{Short Term Time Series Forecasting Performance of AR-HDC \& Seq2Seq-HDC compared to the baseline. We report the mean of RSE and CORR of the experiments. The results in bold are the best and in blue and underlined are second best}
\label{short-term}
\vspace{-0.5cm}
\end{table*}

\section{Experiments}\label{exp}
\subsection{Experimental settings}
\subsubsection{Datasets \& metrics}
We empirically validate TSF-HD on eight real-world benchmark datasets (ETTh1 and ETTh2 (hourly electric transformer data), ETTm1 and ETTm2 (minute-by-minute electric transformer data), WTH (Weather forecasting), ECL (hourly electricity consumption), Exchange (currency exchange rates) and ILI (Influenza-like illness occurrence)), the details of which are in Appendix \ref{dataset_details}. For short-term Time Series Forecasting (TSF), all eight datasets were utilized for evaluation. For long-term TSF, all datasets except ETTh2 and ETTm2 were used. We also use a synthetic abrupt dataset (called \textit{S-A}), derived from \cite{pham2022learning}, to examine speed of adaptation to time series task shifts. This univariate dataset ($N=1$) contains abrupt and recurrent components, where samples switch between different autoregressive (AR) processes.

To evaluate model precision we use the Root Relative Squared Error (RSE) and Empirical Correlation Coefficient (CORR) metrics, following \cite{lai2018modeling}. Details of these metrics are available in Appendix \ref{metrics_appendix}.
To evaluate model overhead and efficiency we record inference latency in seconds and power use (on edge platforms) in watts (W).

\subsubsection{Model baselines \& Implementation Setup}
We compare TSF-HD to several online learning baselines (\textit{FSNet} \cite{pham2022learning}, \textit{ER}\cite{chaudhry2019tiny}, \textit{DER++} \cite{buzzega2020dark}, \textit{OnlineTCN} \cite{woo2022cost}), transformers ( \textit{Informer} \cite{zhou2021informer}), convolutional TSF models (\textit{SCINet} \cite{liu2022scinet}), linear TSF models (\textit{NLinear} \cite{zeng2023transformers} ), a naive direct multi-step that repeats the value of the look-back window (\textit{Repeat}) and Gradient Boosting Regression Trees (\textit{GBRT} \cite{friedman2001greedy}). Further details of the baselines can be found in Appendix \ref{baselines_appendix}. For brevity, in the main body of the paper we present the online learning and transformer baselines. TSF-HD sees similar effectiveness when compared to the other baselines, as shown in Appendix \ref{power_results}.

Similar to prior work \cite{pham2022learning}, each time-series forecasting scenario is split into: (1) A \textit{warmup} phase where the model learns from a small portion of the dataset without measuring model performance, and (2) An \textit{online learning} phase where model predictions are evaluated and the model parameters are updated using the ground truth data as it is made available. In all scenarios, the trainable encoder weight matrix $W_e$ and bias $b_e$ and the trainable regressor hypervectors $W_r$ and bias $b_r$ are initially sampled from the uniform distribution with the interval $\left[\frac{-1}{T},\frac{1}{T}\right]$ before training begins. 
TSF-HD is trained periodically, predicting the next $\tau$ samples and then updating model weights when the system reveals those points, ensuring no overlap between predictions. NLinear, GBRT, Informer and SCINet are not meant to be trained online. Following \cite{pham2022learning}, we thus trained them only in the warm-up phase for 10 epochs. For these baselines, the training/validation set proportion is 75:25. For the online learning baselines the warm-up/online learning phase proportion is 25:75. 

We fix the hypervector dimension at $D=1000$. The look-back window is fixed to twice the forecast horizon: $T = 2\tau$. For short term TSF, $\tau \in \{ 3, 6, 12\}$ and for the long term TSF, $\tau \in \{ 96, 192, 384\}$ for all datasets except ILI where $\tau \in \{ 24, 36, 48\}$ and Exchange where $\tau \in \{ 96, 192, 224\}$. The experiments are run five times with different random seed values, with the mean RSE and CORR reported. 

\begin{table*}[ht]
\begin{center}
\begin{adjustbox}{center, width=0.75\textwidth}
\begin{tabular}{cc|cccccccccccccc|}
\cline{3-16}
 &
   &
  \multicolumn{2}{c}{{\color[HTML]{5B277D} \textbf{Seq2Seq-HDC}}} &
  \multicolumn{2}{c}{{\color[HTML]{5B277D} \textbf{AR-HDC}}} &
  \multicolumn{2}{c}{{\color[HTML]{ACB20C} \textbf{Fsnet}}} &
  \multicolumn{2}{c}{{\color[HTML]{ACB20C} \textbf{ER}}} &
  \multicolumn{2}{c}{{\color[HTML]{ACB20C} \textbf{DR++}}} &
  \multicolumn{2}{c}{{\color[HTML]{ACB20C} \textbf{OnlineTCN}}} &
  \multicolumn{2}{c|}{{\color[HTML]{2A6099} \textbf{Informer}}} \\ \cline{2-16} 
\multicolumn{1}{c|}{} &
  \textit{$\tau$} &
  \textit{\textbf{RSE}} &
  \textit{\textbf{CORR}} &
  \textit{\textbf{RSE}} &
  \textit{\textbf{CORR}} &
  \textit{\textbf{RSE}} &
  \textbf{CORR} &
  \textit{\textbf{RSE}} &
  \textit{\textbf{CORR}} &
  \textit{\textbf{RSE}} &
  \textit{\textbf{CORR}} &
  \textit{\textbf{RSE}} &
  \textit{\textbf{CORR}} &
  \textit{\textbf{RSE}} &
  \textit{\textbf{CORR}} \\ \hline
\multicolumn{1}{|c|}{} &
  96 &
  {\color[HTML]{3465A4} {\underline{0.601}}} &
  \textbf{0.846} &
  \textbf{0.568} &
  {\color[HTML]{3465A4} {\underline{0.84}}} &
  1.073 &
  0.722 &
  0.864 &
  0.778 &
  0.871 &
  0.778 &
  0.897 &
  0.763 &
  1.397 &
  0.629 \\
\multicolumn{1}{|c|}{} &
  192 &
  {\color[HTML]{3465A4} {\underline{0.678}}} &
  {\color[HTML]{3465A4} {\underline{0.803}}} &
  \textbf{0.599} &
  \textbf{0.827} &
  1.353 &
  0.626 &
  1.089 &
  0.696 &
  1.066 &
  0.703 &
  1.118 &
  0.675 &
  1.455 &
  0.627 \\
\multicolumn{1}{|c|}{\multirow{-3}{*}{\textbf{ETTh1}}} &
  384 &
  {\color[HTML]{3465A4} {\underline{0.797}}} &
  {\color[HTML]{3465A4} {\underline{0.739}}} &
  \textbf{0.628} &
  \textbf{0.831} &
  1.228 &
  0.616 &
  1.347 &
  0.609 &
  1.295 &
  0.617 &
  1.347 &
  0.604 &
  1.508 &
  0.625 \\ \hline
\multicolumn{1}{|c|}{} &
  96 &
  {\color[HTML]{3465A4} {\underline{0.315}}} &
  0.951 &
  \textbf{0.297} &
  \textbf{0.965} &
  0.355 &
  0.951 &
  0.347 &
  0.953 &
  0.341 &
  {\color[HTML]{3465A4} {\underline{0.956}}} &
  0.359 &
  0.951 &
  0.822 &
  0.856 \\
\multicolumn{1}{|c|}{} &
  192 &
  0.392 &
  0.925 &
  {\color[HTML]{3465A4} {\underline{0.373}}} &
  \textbf{0.949} &
  0.409 &
  0.937 &
  0.374 &
  0.942 &
  \textbf{0.372} &
  {\color[HTML]{3465A4} {\underline{0.943}}} &
  0.376 &
  0.941 &
  0.855 &
  0.862 \\
\multicolumn{1}{|c|}{\multirow{-3}{*}{\textbf{ETTm1}}} &
  384 &
  0.465 &
  0.893 &
  \textbf{0.431} &
  0.924 &
  0.522 &
  0.915 &
  0.454 &
  {\color[HTML]{3465A4} {\underline{0.926}}} &
  {\color[HTML]{3465A4} {\underline{0.448}}} &
  \textbf{0.927} &
  0.476 &
  0.925 &
  0.841 &
  0.859 \\ \hline
\multicolumn{1}{|c|}{} &
  24 &
  {\color[HTML]{3465A4} {\underline{0.288}}} &
  0.957 &
  \textbf{0.229} &
  \textbf{0.986} &
  0.304 &
  {\color[HTML]{3465A4} {\underline{0.967}}} &
  0.382 &
  0.936 &
  0.357 &
  0.947 &
  0.411 &
  0.932 &
  1.09 &
  0.211 \\
\multicolumn{1}{|c|}{} &
  36 &
  0.391 &
  0.92 &
  \textbf{0.199} &
  \textbf{0.99} &
  {\color[HTML]{3465A4} {\underline{0.389}}} &
  {\color[HTML]{3465A4} {\underline{0.945}}} &
  0.454 &
  0.931 &
  \textit{0.489} &
  \textit{0.926} &
  0.477 &
  0.927 &
  1.122 &
  0.221 \\
\multicolumn{1}{|c|}{\multirow{-3}{*}{\textbf{ILI}}} &
  48 &
  {\color[HTML]{3465A4} {\underline{0.396}}} &
  {\color[HTML]{3465A4} {\underline{0.919}}} &
  \textbf{0.204} &
  \textbf{0.989} &
  0.503 &
  0.916 &
  0.534 &
  0.893 &
  0.553 &
  0.895 &
  0.582 &
  0.879 &
  0.193 &
  0.144 \\ \hline
\multicolumn{1}{|c|}{} &
  96 &
  {\color[HTML]{3465A4} {\underline{0.211}}} &
  {\color[HTML]{3465A4} {\underline{0.978}}} &
  \textbf{0.097} &
  \textbf{0.997} &
  2.83 &
  0.373 &
  1.353 &
  0.586 &
  1.154 &
  0.641 &
  2.241 &
  0.395 &
  0.904 &
  0.577 \\
\multicolumn{1}{|c|}{} &
  192 &
  {\color[HTML]{3465A4} {\underline{0.318}}} &
  {\color[HTML]{3465A4} {\underline{0.948}}} &
  \textbf{0.155} &
  \textbf{0.994} &
  2.832 &
  0.254 &
  2.311 &
  0.252 &
  2.19 &
  \textit{0.254} &
  2.993 &
  0.203 &
  0.968 &
  0.398 \\
\multicolumn{1}{|c|}{\multirow{-3}{*}{\textbf{Exchange}}} &
  224 &
  {\color[HTML]{3465A4} {\underline{0.328}}} &
  {\color[HTML]{3465A4} {\underline{0.945}}} &
  \textbf{0.137} &
  \textbf{0.995} &
  3.697 &
  0.162 &
  2.204 &
  0.261 &
  2.106 &
  0.265 &
  2.436 &
  0.215 &
  0.989 &
  0.335 \\ \hline
\multicolumn{1}{|c|}{} &
  96 &
  \textbf{0.177} &
  {\color[HTML]{3465A4} {\underline{0.985}}} &
  {\color[HTML]{3465A4} {\underline{0.206}}} &
  \textbf{0.989} &
  0.388 &
  0.947 &
  0.221 &
  0.982 &
  0.215 &
  0.982 &
  0.219 &
  0.978 &
  1 &
  0.027 \\
\multicolumn{1}{|c|}{} &
  192 &
  \textbf{0.214} &
  {\color[HTML]{3465A4} {\underline{0.977}}} &
  {\color[HTML]{3465A4} {\underline{0.294}}} &
  \textbf{0.979} &
  0.413 &
  0.913 &
  0.303 &
  0.952 &
  0.306 &
  0.951 &
  0.305 &
  0.951 &
  1 &
  0.031 \\
\multicolumn{1}{|c|}{\multirow{-3}{*}{\textbf{ECL}}} &
  384 &
  \textbf{0.282} &
  {\color[HTML]{3465A4} {\underline{0.962}}} &
  {\color[HTML]{3465A4} {\underline{0.302}}} &
  \textbf{0.979} &
  0.517 &
  0.861 &
  0.504 &
  0.866 &
  0.484 &
  0.866 &
  0.521 &
  0.857 &
  1 &
  0.077 \\ \hline
\multicolumn{1}{|c|}{} &
  96 &
  {\color[HTML]{3465A4} {\underline{0.697}}} &
  {\color[HTML]{3465A4} {\underline{0.793}}} &
  \textbf{0.633} &
  \textbf{0.796} &
  0.886 &
  0.766 &
  0.807 &
  0.787 &
  0.798 &
  0.789 &
  0.838 &
  0.782 &
  0.841 &
  0.567 \\
\multicolumn{1}{|c|}{} &
  192 &
  {\color[HTML]{3465A4} {\underline{0.723}}} &
  {\color[HTML]{3465A4} {\underline{0.784}}} &
  \textbf{0.632} &
  \textbf{0.791} &
  0.911 &
  0.753 &
  0.854 &
  0.771 &
  0.845 &
  0.772 &
  0.926 &
  0.759 &
  0.895 &
  0.549 \\
\multicolumn{1}{|c|}{\multirow{-3}{*}{\textbf{WTH}}} &
  384 &
  {\color[HTML]{3465A4} {\underline{0.777}}} &
  {\color[HTML]{3465A4} {\underline{0.762}}} &
  \textbf{0.653} &
  \textbf{0.776} &
  0.965 &
  0.726 &
  0.936 &
  0.743 &
  0.894 &
  0.749 &
  1.014 &
  0.732 &
  0.871 &
  0.547 \\ \hline
\end{tabular}
\end{adjustbox}
\end{center}
\caption{Long Term Time Series Forecasting Performance of AR-HDC \& Seq2Seq-HDC compared to the baseline. We report the mean of RSE and CORR of the experiments. The results in bold are the best and in blue and underlined are second best}
\label{long-term}
\vspace{-0.25cm}
\end{table*}
\begin{figure*}[htp]
    \begin{subfigure}[b]{\textwidth}
        \centering
        \includegraphics[width=0.8\textwidth]{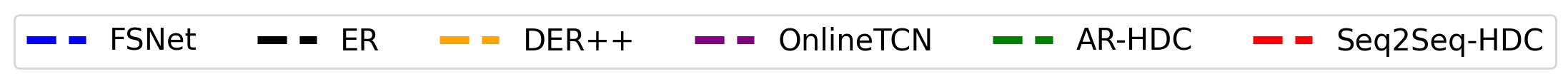}
        \caption*{} 
        \label{fig:legend}
        \vspace{-0.5cm}
    \end{subfigure}

    \centering
    \begin{subfigure}[b]{0.3\textwidth}
        \centering
        \includegraphics[width=\textwidth]{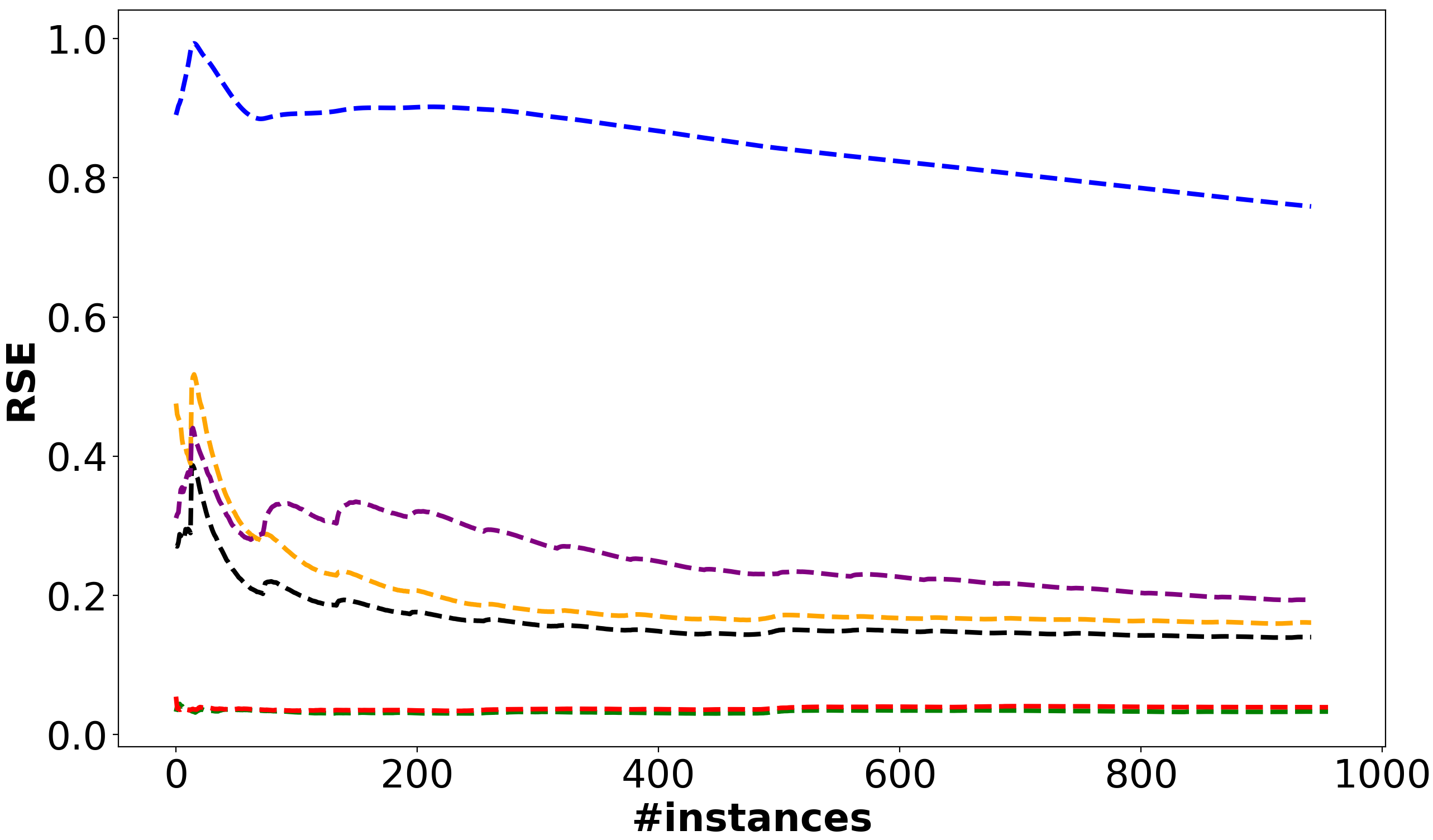}
        \caption{Exchange}
        \label{fig:sub1}
    \end{subfigure}
    \hfill
    \begin{subfigure}[b]{0.3\textwidth}
        \centering
        \includegraphics[width=\textwidth]{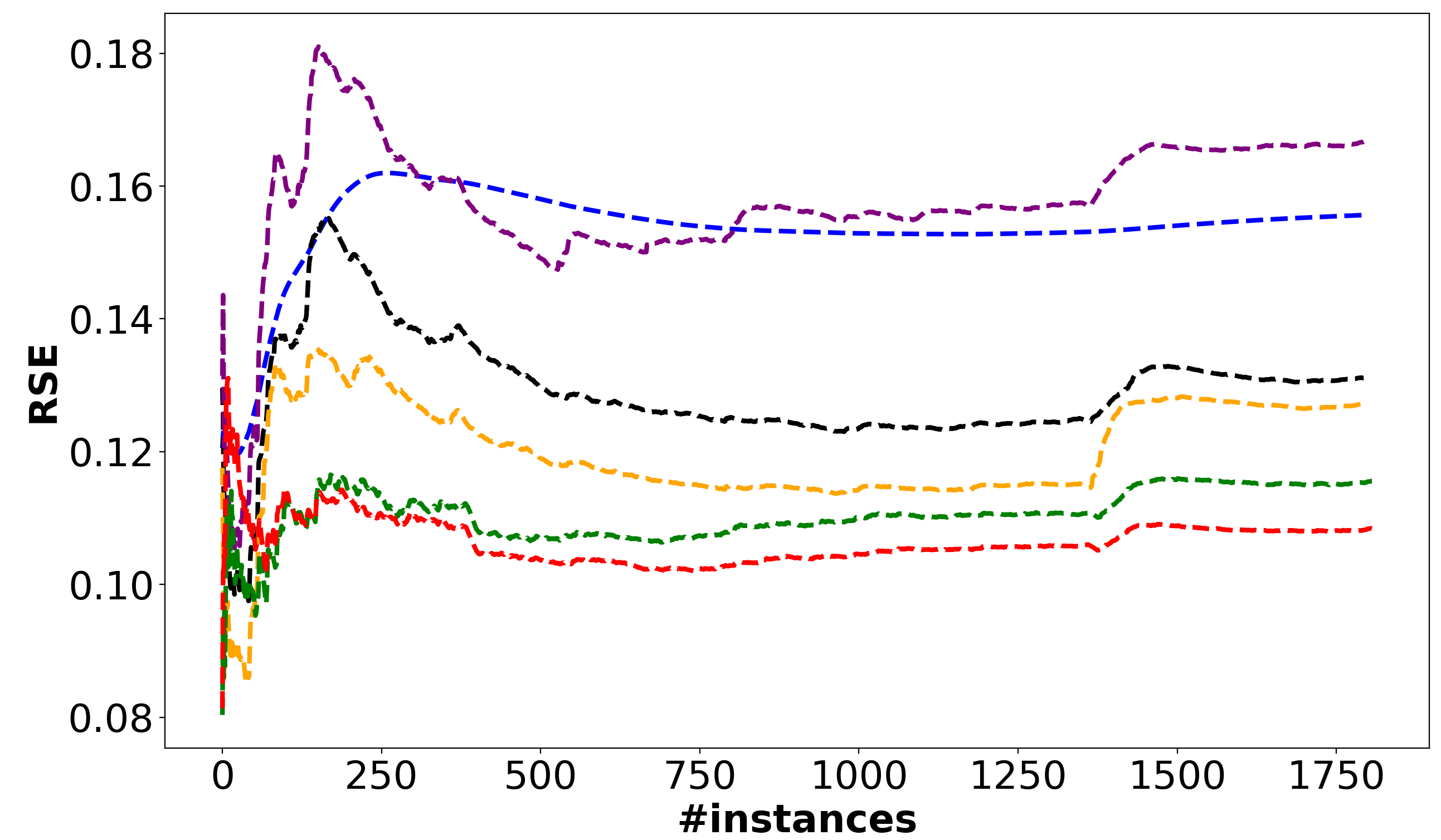}
        \caption{ETTm2}
        \label{fig:sub2}
    \end{subfigure}
    \hfill
    \begin{subfigure}[b]{0.3\textwidth}
        \centering
        \includegraphics[width=\textwidth]{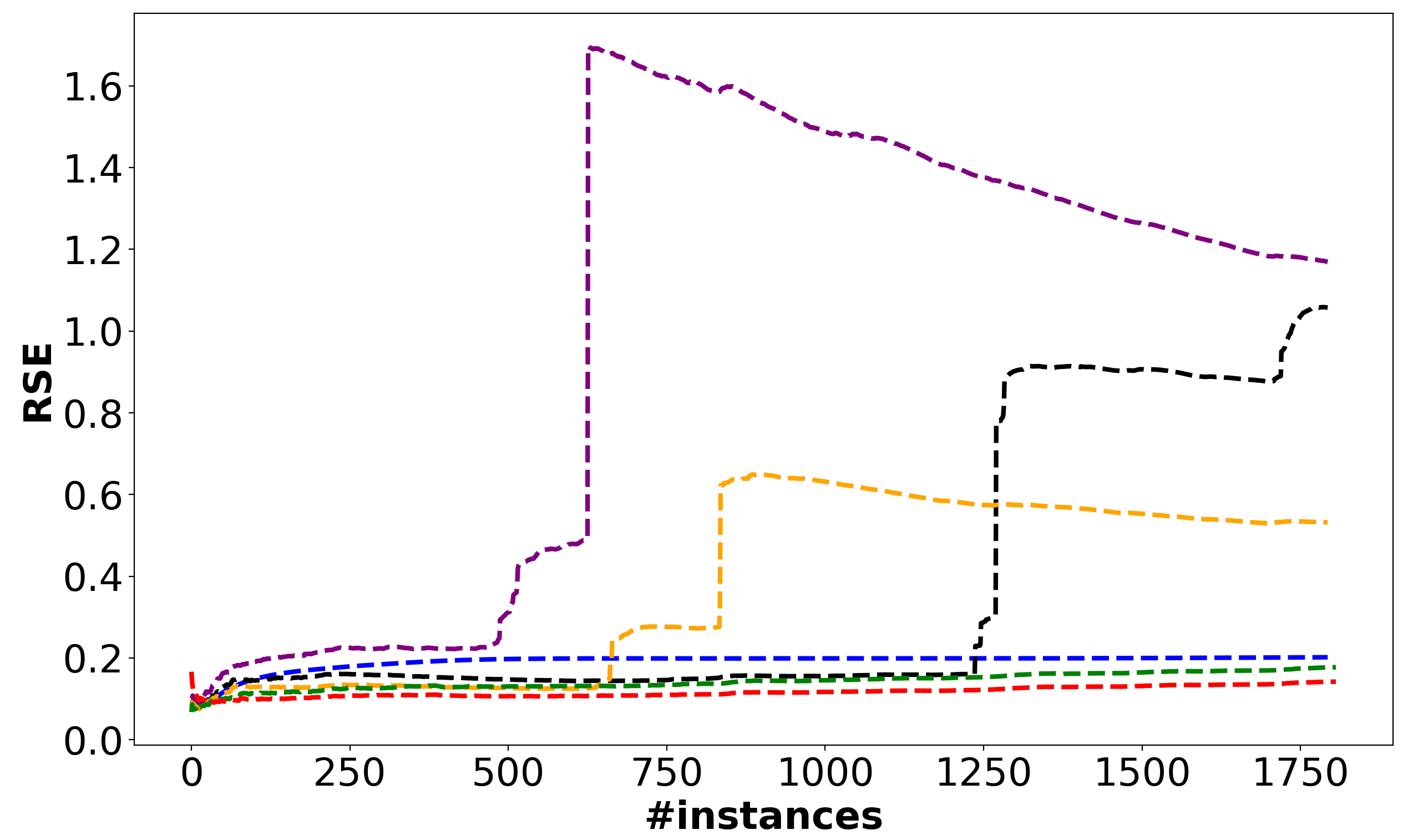}
        \caption{ETTm1}
        \label{fig:sub3}
    \end{subfigure}


    \caption{Evolution of the cumulative RSE during online learning with forecasting window $\tau=6$}
    \label{cvg}
    \vspace{-0.5cm}
\end{figure*}

\subsection{Performance Analysis}
\subsubsection{Model Precision} \label{precision}
Table \ref{short-term} shows the RSE and CORR for the two TSF-HD frameworks (AR-HDC and Seq2Seq-HDC) compared to the baselines. For short-term TSF, AR-HDC surpasses all state-of-the-art (SOTA) models all 24 test cases with either lower RSE and higher CORR. 
Seq2Seq-HDC shows comparable performance, outshining SOTA models in 10 out of 24 test cases, and is the top algorithm in 2 of these cases. We thus see that for a \textit{majority} of the test cases in short-term TSF, TSF-HD frameworks outperform the state of the art.

Table \ref{long-term} presents validation results for long-term TSF, comparing the TSF-HD frameworks to the state of the art. We see that both TSF-HD frameworks perform better in comparison to the state of the art for the long-term as opposed to the short-term TSF cases. AR-HD outperforms the baseline models in either CORR or RSE in 16 out of 18 test cases. Seq2Seq-HDC achieves the best results in 4 out of 18 cases for RSE and ranks second in 10 of the 18 cases. Notably, AR-HDC demonstrates impressive performance in the ETTh1, ETTm1, ILI, and WTH datasets for very long forecast horizons ($\tau=384$). Due to the high value of $D$ compared to $\tau$, TSF-HD frameworks are thus able to perform accurate hyperdimensional linear regression after encoding the input data, leading to this high performance.

The high values of RSE seen for the baselines in Tables \ref{short-term} and \ref{long-term} (and for offline predictor data in Appendix \ref{power_results}), particularly for long-term TSF, indicate a collapse in performance compared to a naive predictor. This is in part due to increased values of $\tau$ and in part due to the fact that our experiments use raw data without standardization or normalization, since in an online learning scenario with shifting data streams we do not know the mean and standard deviation of input data.

\begin{figure*}[htp]
    \begin{subfigure}[b]{0.8\textwidth}
        \centering
        \includegraphics[width=\textwidth]{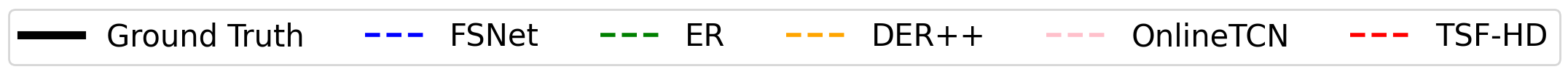}
        \label{fig:sub1}
        \vspace{-0.5cm}
    \end{subfigure}
    \centering
    \begin{subfigure}[b]{0.24\textwidth}
        \centering
        \includegraphics[width=\textwidth]{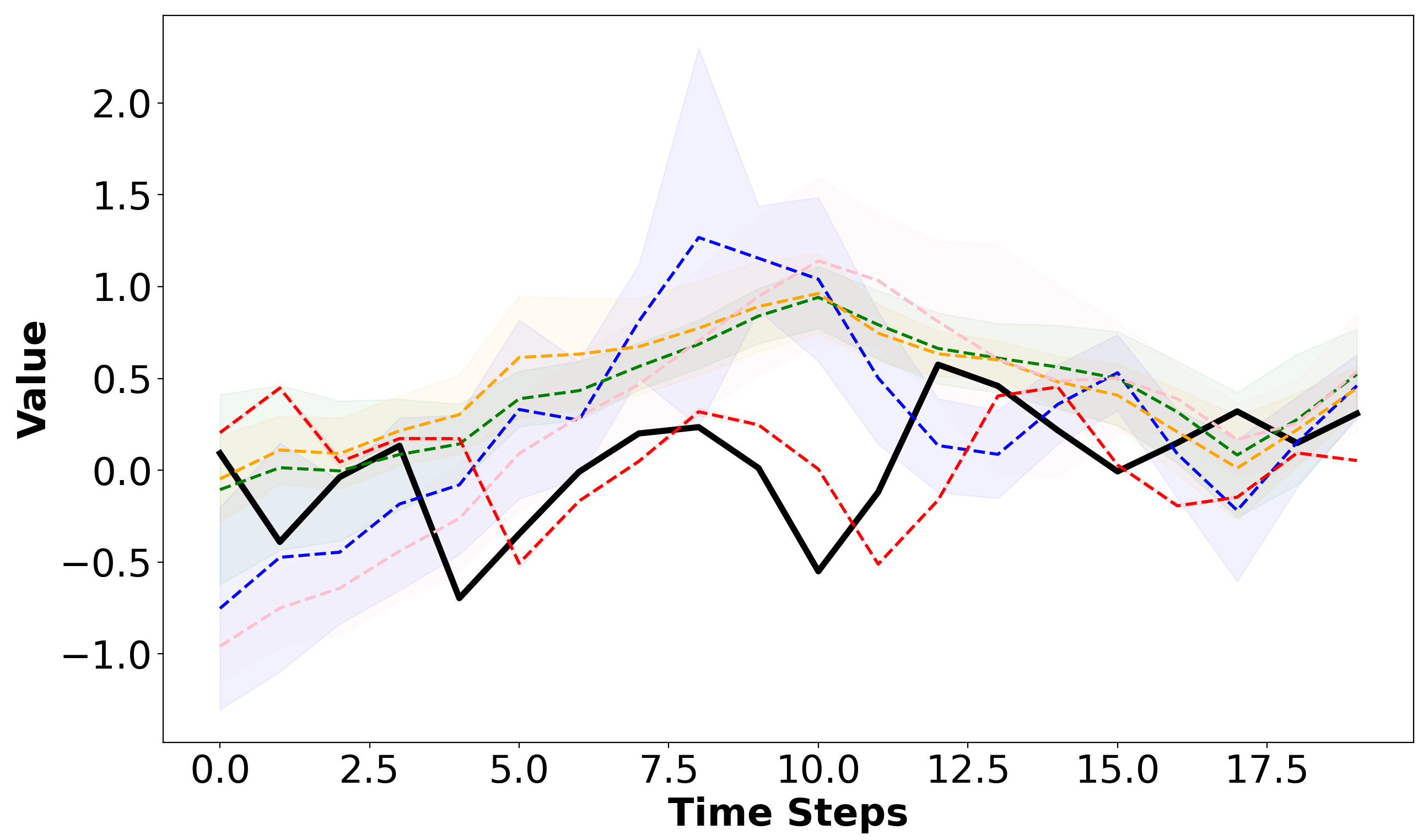}
        \caption{$1^{st}$ task shift zone}
        \label{fig:sub1}
    \end{subfigure}
    \hfill
    \begin{subfigure}[b]{0.24\textwidth}
        \centering
        \includegraphics[width=\textwidth]{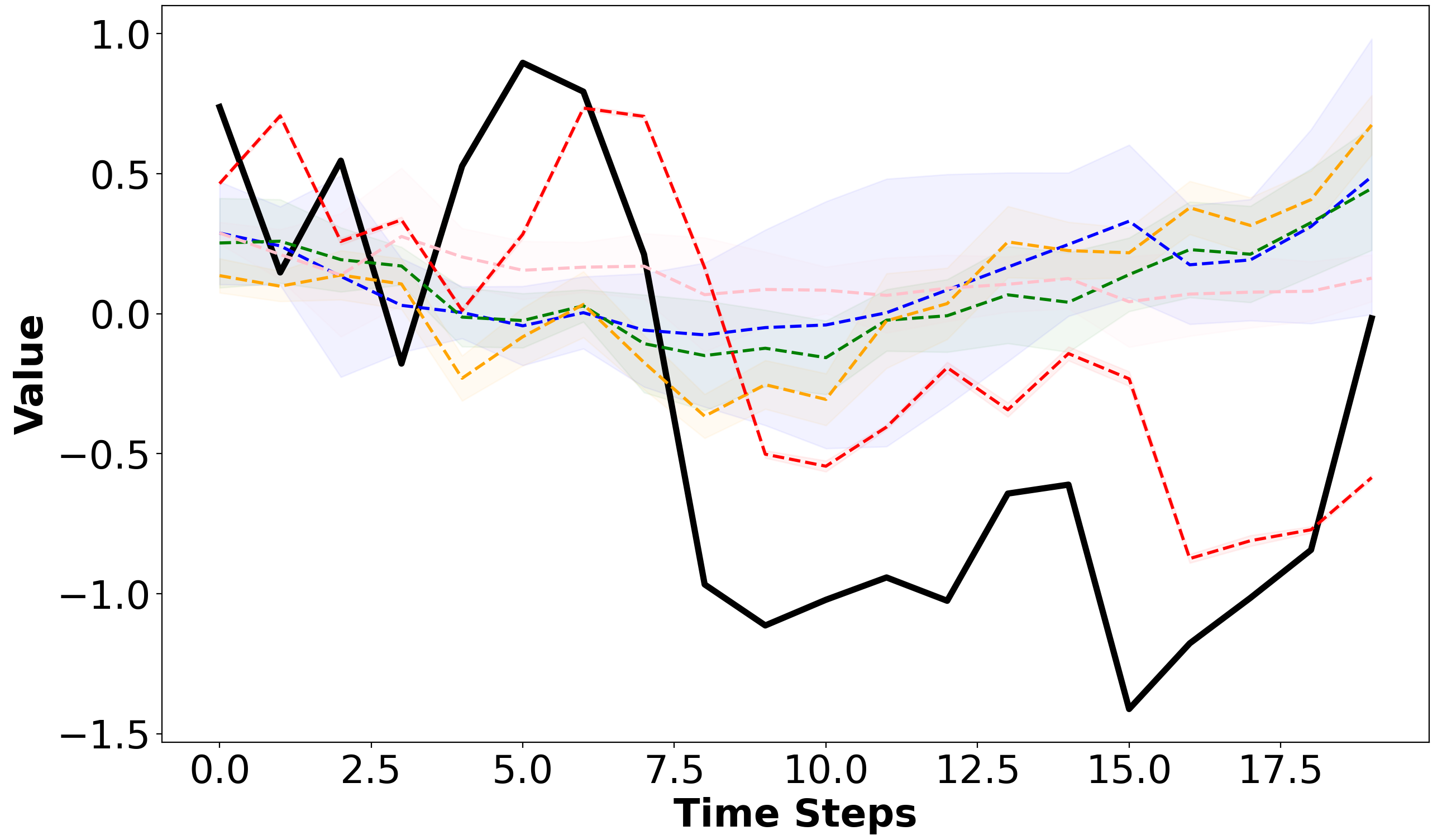}
        \caption{$2^{nd}$ task shift zone}
        \label{fig:sub2}
    \end{subfigure}
    \hfill
    \begin{subfigure}[b]{0.24\textwidth}
        \centering
        \includegraphics[width=\textwidth]{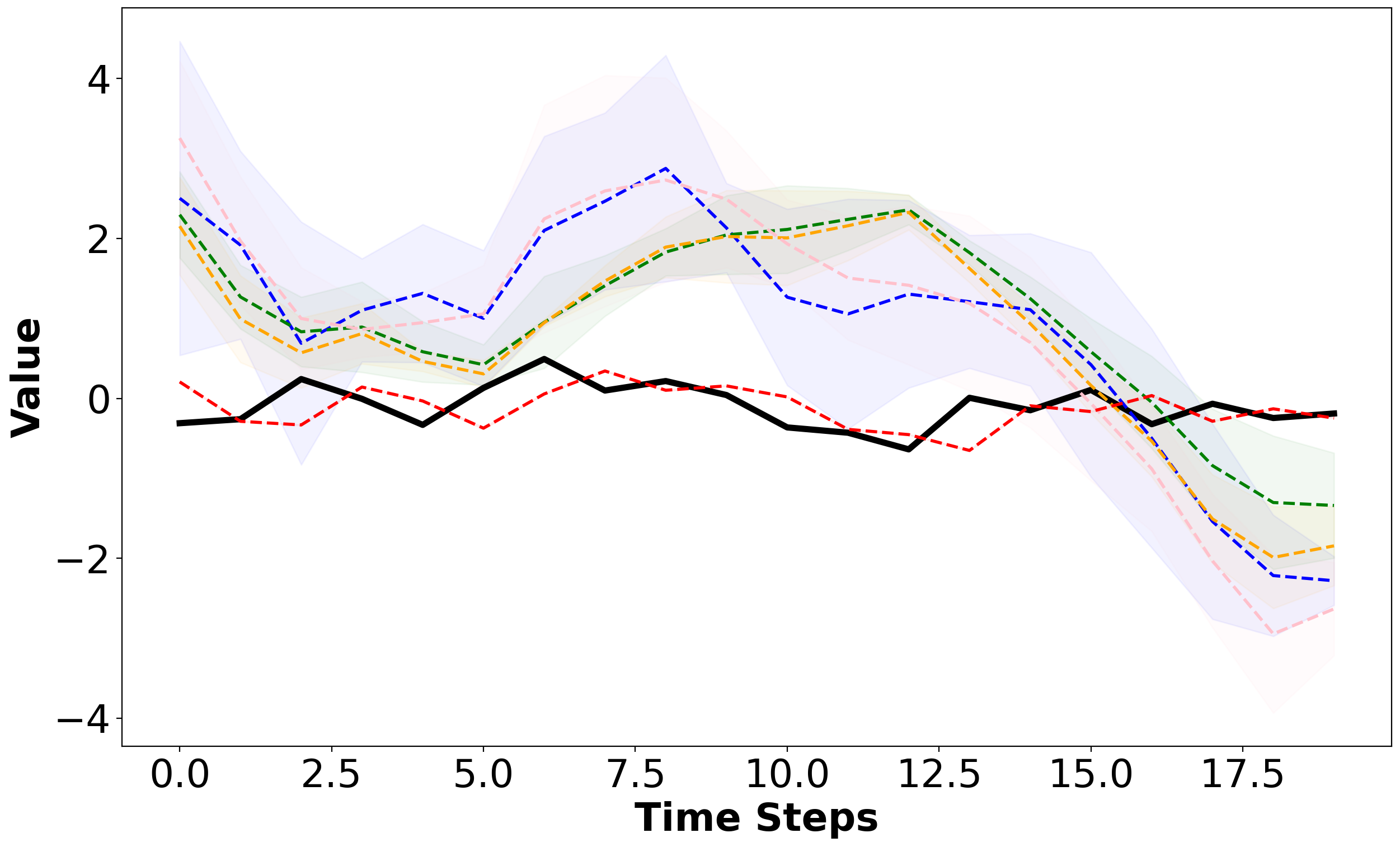}
        \caption{$3^{rd}$ task shift zone}
        \label{fig:sub3}
    \end{subfigure}
    \hfill
    \begin{subfigure}[b]{0.24\textwidth}
        \centering
        \includegraphics[width=\textwidth]{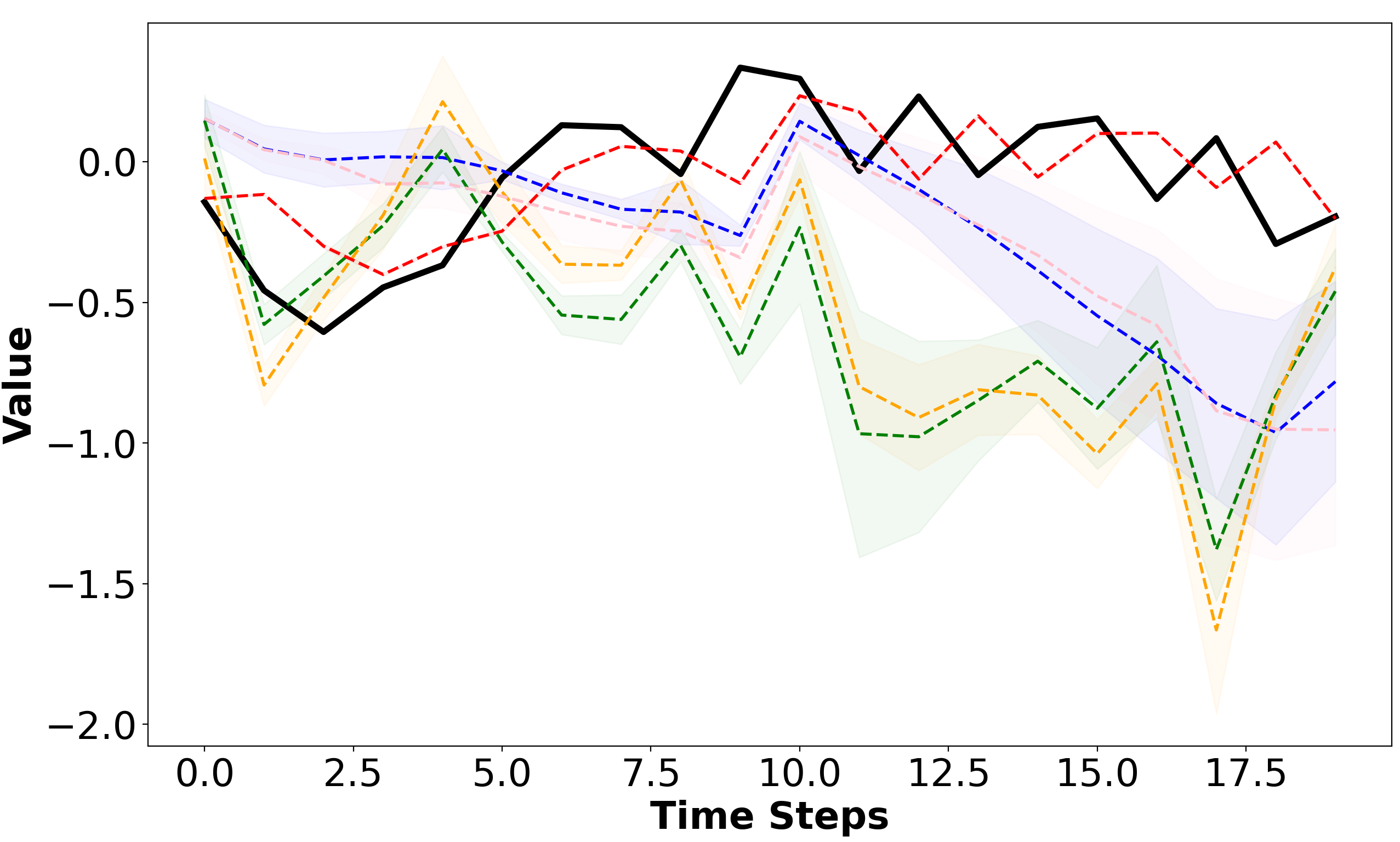}
        \caption{$4^{th}$ task shift zone}
        \label{fig:sub4}
    \end{subfigure}
    \caption{Visualization of the baseline and TSF-HD method performance as the tasks shift between four different autoregressive processes.}
    \label{viz}
    \vspace{-0.25cm}
\end{figure*}
\begin{table*}[ht]
\centering
\begin{adjustbox}{width=0.8\textwidth, center}
\begin{tabular}{cc|cccc|cccc|}
\cline{3-10}
 &
   &
  \multicolumn{4}{c|}{\textit{Latency}} &
  \multicolumn{4}{c|}{\textit{Power}} \\ \cline{2-10} 
\multicolumn{1}{c|}{} &
  \textit{$\tau$} &
  \textbf{\color[HTML]{5B277D} AR-HDC} &
  \textbf{\color[HTML]{5B277D} Seq2Seq-HDC} &
  \textbf{\color[HTML]{ACB20C} FSNet} &
  \textbf{\color[HTML]{ACB20C} OnlineTCN} &
  \textbf{\color[HTML]{5B277D} AR-HDC} &
  \textbf{\color[HTML]{5B277D} Seq2Seq-HDC} &
  \textbf{\color[HTML]{ACB20C} FSNet} &
  \textbf{\color[HTML]{ACB20C} OnlineTCN} \\ \hline
\multicolumn{1}{|c|}{} &
  3 &
   $\textbf{0.016}^{\pm \textbf{0.002}}$ &
  \color[HTML]{2A6099}\underline{ $0.066^{\pm 0.013}$} &
  $0.452^{\pm 0.121}$ &
  $0.213^{\pm 0.079}$ &
  $\textbf{3.395}^{\pm \textbf{0.534}}$ &
  \color[HTML]{2A6099}\underline{ $4.246^{\pm 0.945}$} &
  $4.855^{\pm 0.635}$ &
  $4.654^{\pm 0.748}$ \\
\multicolumn{1}{|c|}{\multirow{-2}{*}{\textbf{Exchange}}} &
  96 &
  $0.937^{\pm 0.347}$ &
  $\textbf{0.073}^{\pm \textbf{0.041}}$ &
  $0.538^{\pm 0.227}$ &
  \color[HTML]{2A6099}\underline{ $0.273^{\pm 0.093}$} &
  $5.257^{\pm 0.511}$ &
  $\textbf{4.437}^{\pm \textbf{0.855}}$ &
  $4.975^{\pm 0.642}$ &
  \color[HTML]{3465A4}\underline{ $4.788^{\pm 0.804}$} \\ \hline
\multicolumn{1}{|c|}{} &
  3 &
  $\textbf{0.0144}^{\pm \textbf{0.001}}$ &
  \color[HTML]{2A6099}\underline{ $0.065^{\pm 0.011}$} &
  $0.47^{\pm 0.122}$ &
  $0.212^{\pm 0.078}$ &
  $\textbf{3.508}^{\pm \textbf{0.486}}$ &
  \color[HTML]{3465A4}\underline{ $4.236^{\pm 0.904}$} &
  $4.766^{\pm 0.637}$ &
  $4.567^{\pm 0.783}$ \\
\multicolumn{1}{|c|}{\multirow{-2}{*}{\textbf{ETTh1}}} &
  96 &
  $0.867^{\pm 0.03}$ &
  $\textbf{0.071}^{\pm \textbf{0.038}}$ &
  $0.530^{\pm 0.209}$ &
  \color[HTML]{3465A4}\underline{ $0.295^{\pm 0.125}$} &
  $5.286^{\pm 0.203}$ &
  $\textbf{4.419}^{\pm \textbf{0.967}}$ &
  $4.961^{\pm 0.591}$ &
  \color[HTML]{3465A4}\underline{ $4.858^{\pm 0.698}$} \\ \hline
\end{tabular}
\end{adjustbox}
\caption{Power consumption \& latency of two online learning baselines and the TSF-HD model on RaspberryPI ($mean^{\pm std}$)}
\label{power-latency-cpu-paper}
\vspace{-0.5cm}
\end{table*}
\begin{table}[ht]
\begin{center}
\begin{adjustbox}{width=0.9\columnwidth}
\begin{tabular}{ccc|cccc}
\hline
\multicolumn{3}{c|}{\textbf{Dimension}}               & \textbf{0.5k} & \textbf{1k} & \textbf{5k} & \textbf{10k} \\ \hline
\multicolumn{1}{c|}{\multirow{4}{*}{\textbf{RSE}}} & \multicolumn{1}{c|}{\multirow{2}{*}{\textbf{Seq2Seq-HDC}}} & ETTh2 & 0.971 & 0.756 & 0.164 & 0.143 \\
\multicolumn{1}{c|}{} & \multicolumn{1}{c|}{} & ETTm2 & 0.849         & 0.349       & 0.179       & 0.178        \\ \cline{2-7} 
\multicolumn{1}{c|}{}                              & \multicolumn{1}{c|}{\multirow{2}{*}{\textbf{AR-HDC}}}      & ETTh2 & 0.059 & 0.059 & 0.059 & 0.058 \\
\multicolumn{1}{c|}{} & \multicolumn{1}{c|}{} & ETTm2 & 0.047         & 0.049       & 0.052       & 0.053        \\ \hline
\end{tabular}
\end{adjustbox}
\caption{\centering Effect of hypervector dimensionality reduction on the RSE of ETTh2 \& ETTm2 datasets TSF with $\tau=96$}
\label{dimension}
\end{center}
\vspace{-0.5cm}
\end{table}

\subsubsection{Convergence of TSF-HD Models}
Figure \ref{cvg} shows the cumulative average RSE of the online learning baselines for short-term TSF ($\tau = 6$) compared to AR-HDC and Seq2Seq-HDC on the Exchange, ETTm1 and ETTm2 datasets. 

We see that in the first $20\%$ of the data (Exchange \& ETTm2), all models suffer from concept drift and attempt to adapt to the shift in trend. The remainder of the data appears stationary, as indicated by the relative convergence of the RSE for all test cases. 
For all test cases, AR-HDC outperforms all other models, achieving faster RSE convergence (quicker adaptation to task shifts) and maintaining a generally lower RSE value (efficient adaptation).

The Seq2Seq-HDC model's RSE convergence follows a similar trend to AR-HDC for ETTm1 and Exchange. For all the test cases, the performance (i.e, speed of convergence and asymptotic value) of AR-HDC and Seq2Seq-HDC are comparable except for ETTm2, where Seq2Seq-HDC a has lower final RSE value. In the ETTm1 case, OnlineTCN, ER and DER++ are highly sensitive to shifts. These observations validate our framing of the TSF problem as one of online linear hyperdimensional forecasting, using a trainable mapping from low (input) dimensions to the hyperspace.


\subsubsection{Adaptation to Task Shifts}\label{adaptation}
To analyze the TSF-HD's adaptation to task shifts in data streams, we use the abrupt synthetic dataset S-A \cite{pham2022learning}, composed of different tasks concatenated together. Thus, we begin by running warm-up training on a warm-up autoregressive (AR) process in S-A for 1000 time steps. Following that, we transition to a different AR process, allow the models under evaluation to adapt for 200 time steps, and then record their performance over the next 20 episodes before switching tasks again. Figure \ref{viz} shows the predictions of different online learning models and TSF-HD after conducting the described experiment. We begin with the warmup process $AR1$ before switching to another process $AR2$ (the results of which are shown in Fig. \ref{viz}(a), then $AR1$ again (Fig. \ref{viz}(b), then $AR2$ again (Fig. \ref{viz}(c), then $AR3$ (Fig. \ref{viz}(d)), a new process. Further details on each autoregressive process are discussed in Appendix \ref{dataset_details}. In this experiment, we focused on short-term distribution shift adaptation, with a forecast horizon of $\tau=1$. Both the AR-HDC and Seq2Seq-HDC frameworks are identical for $\tau=1$, and are represented in Figure \ref{viz} as `TSF-HD'. The area around each curve represents one standard deviation around the mean value after repeating the experiment five times. The baselines show more variation and error than TSF-HD, indicating that the linearity of the nonlinear processes in high dimensions enables fast adaptation. 


\subsubsection{Online learning  power and latency efficiency}

We assessed the prediction latency and power usage of TSF-HD and selected online learning baselines on two edge platforms. Our findings for latency and power on two datasets on a Raspberry Pi are detailed in Table \ref{power-latency-cpu-paper}. Similar results are showcased for the NVIDIA Jetson Nano and for further datasets on the Raspberry Pi in Appendix \ref{appendices_hw}. For short-term TSF ($\tau=3$), AR-HDC demonstrates the lowest latency, followed by Seq2Seq-HDC. Notably, both models maintain a nearly constant latency with minimal standard deviation.
For long-term TSF, AR-HDC model falls short, exhibiting the highest latency and reduced power efficiency. This is attributed to its use of a loop for prediction and update phases, updating the model $\tau \gg 1$ times during the update loop. In contrast, Seq2Seq-HDC outperforms the others in both inference latency and power efficiency.

\subsubsection{Effect of Dimensionality reduction}
Table \ref{dimension} illustrates the influence of hypervector dimension $D$ on the Relative Squared Error (RSE) of both Seq2Seq-HDC and AR-HDC models for the ETTh2 and ETTm2 datasets with $\tau=96$. For Seq2Seq-HDC, the RSE decreases by $85\%$ and $79\%$ for ETTh2 and ETTm2 respectively, as $D$ ranges from 500 to 10k, highlighting the advantages of operating in a high-dimensional space. In contrast, an increase in $D$ does not yield similar benefits for AR-HDC due to the greater data efficiency of the autoregressive formulation. 

\section{Conclusion}\label{conc}
In this work, we present \textit{TSF-HD}, a novel online variable-horizon hyperdimensional time-series prediction framework. By reframing the time-series prediction problem as task-free, online hyperdimensional regression, we exploit the linearity of the nonlinear time series in high dimensions, achieving results superior to the state-of-the-art with higher efficiency.


\section*{Acknowledgement}
Acknowledgement removed for review.

\section*{Impact Statement}\label{impact}
This work provides low-overhead solution for edge-based time-series forecasting, potentially enabling consumers and end users to deploy a machine learning forecaster on cheap, rugged hardware with reduced environmental impact from power consumption. While we acknowledge the potential harms from such work, we also note that this provides environmental, economic and accessibility benefits.

One notable concern is the use of more efficient time-series forecasting models for military applications or for unethical financial practices based on information asymmetry. Mitigation strategies to address this may involve increased oversight and monitoring.

Regarding fairness considerations, we believe that the more explainable nature of the hyperdimensional linear regressor allows more thorough research into its bias and fairness. We also note that making sophisticated forecasting capabilities available on cheap hardware and low power platforms contributes to democratizing these capabilities and allowing their use without the expense of complex hardware and high energy use.

\bibliography{example_paper}
\bibliographystyle{icml2024}

\newpage
\appendix
\onecolumn
\section*{Appendices}
\section{Trainable Encoder Validation: Problem Framing}\label{appendix_framing}
\begin{figure}[ht]
  \centering
  \begin{subfigure}[b]{0.32\textwidth}
    \includegraphics[width=\textwidth]{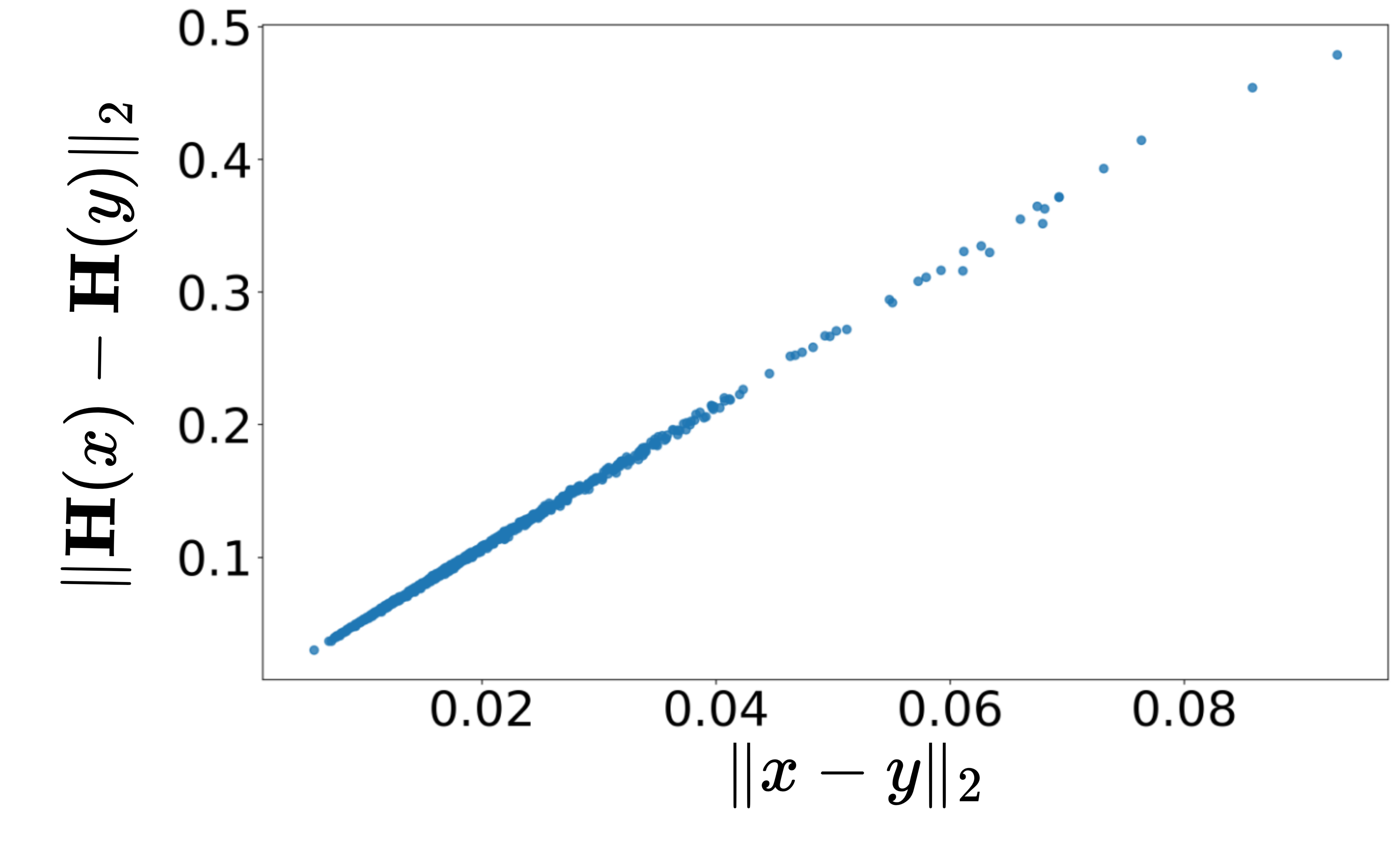}
    \caption{\textit{Exchange}}
    \label{fig:sub1}
  \end{subfigure}
  \hfill 
  \begin{subfigure}[b]{0.32\textwidth}
    \includegraphics[width=\textwidth]{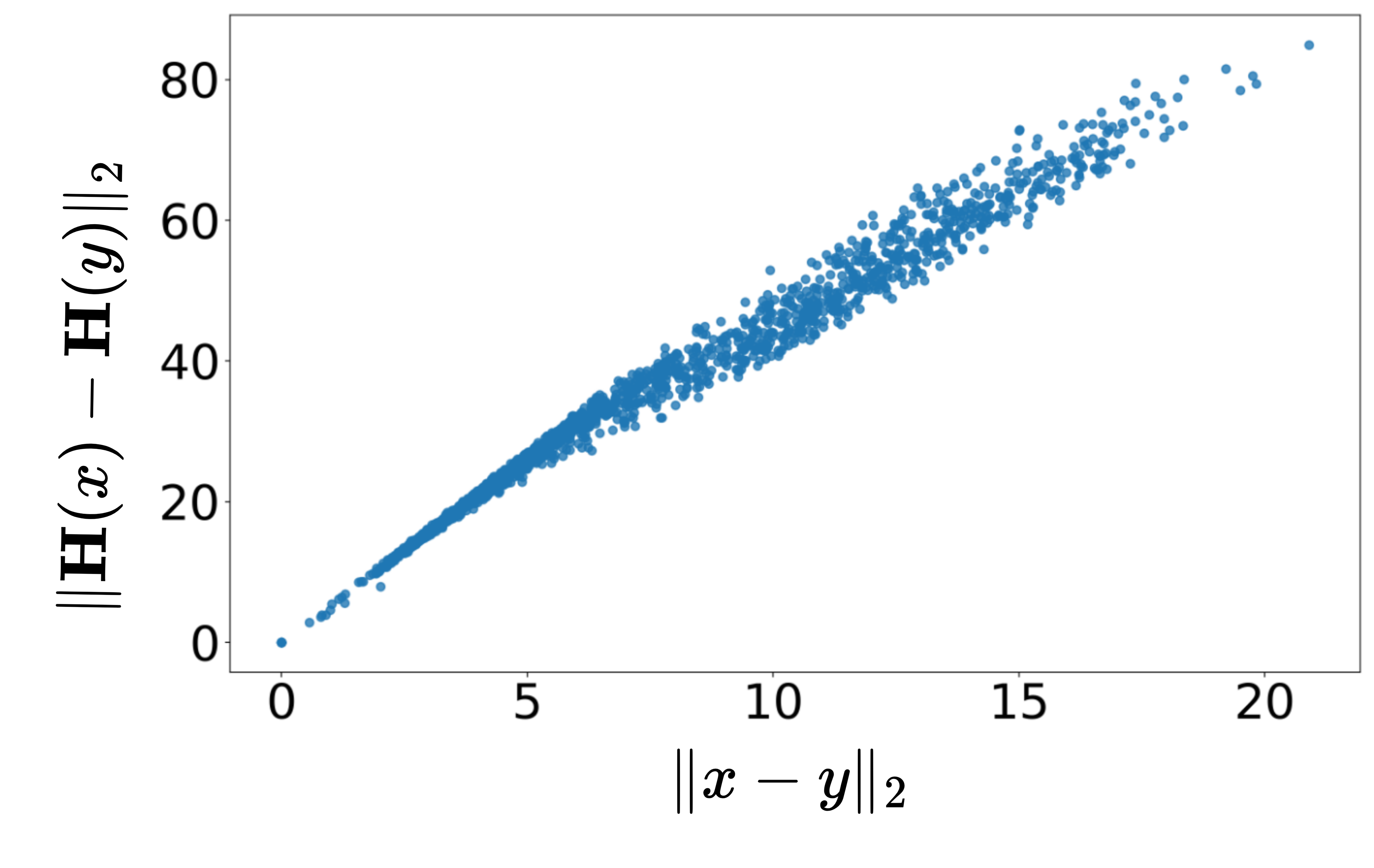}
    \caption{\textit{ETTh1}}
    \label{fig:sub2}
  \end{subfigure}
    \hfill 
  \begin{subfigure}[b]{0.32\textwidth}
    \includegraphics[width=\textwidth]{images/ETTm1_tau_6.png}
    \caption{\textit{ETTm1}}
    \label{fig:sub2}
  \end{subfigure}
  \caption{Distance preservation across hyperdimensional mappings of input sequences $\mathbf{H}(x)$ for $\tau=6$}
  \label{dist_pres_6}
\end{figure}

The figure \ref{dist_pres_6} illustrates the distance preservation of the HDC encoder of AR-HDC and Seq2Seq-HDC for a short term TSF ($\tau=6$) across the Exchange, ETTh1 and the ETTm1 datasets. These readings are similar to the results shown in Figure \ref{distance_1}, with a linear relationship maintained between the distances $\lVert x-y \rVert$ separating two points in the data stream (so that $x,y\in\mathcal{X}$) and the encoded distances $\lVert \mathbf{H}(x)-\mathbf{H}(y)\rVert$. 

\begin{figure}[ht]
  \centering
  \begin{subfigure}[b]{0.32\textwidth}
    \includegraphics[width=\textwidth]{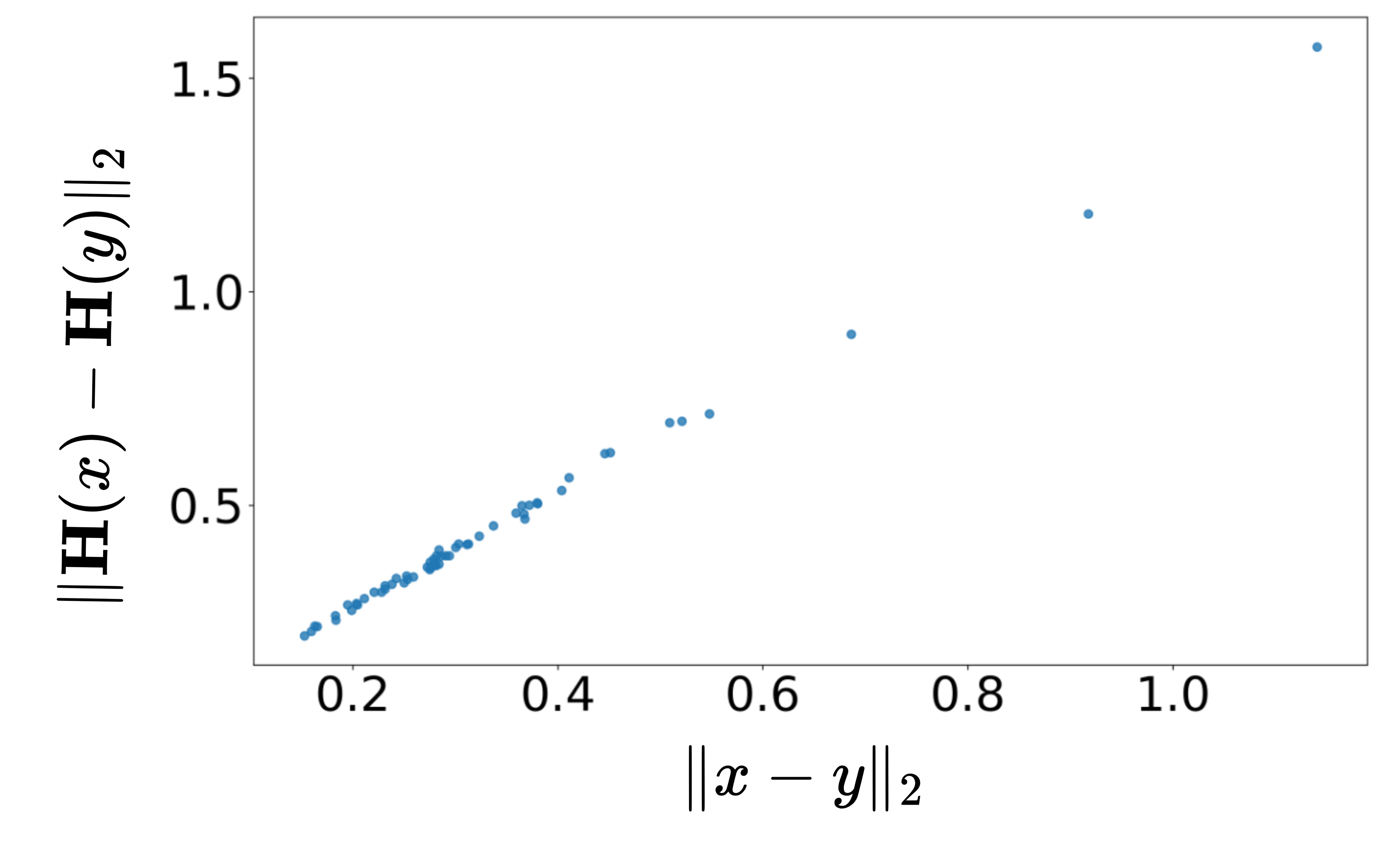}
    \caption{\textit{Exchange}}
    \label{fig:sub1}
  \end{subfigure}
  \hfill 
  \begin{subfigure}[b]{0.32\textwidth}
    \includegraphics[width=\textwidth]{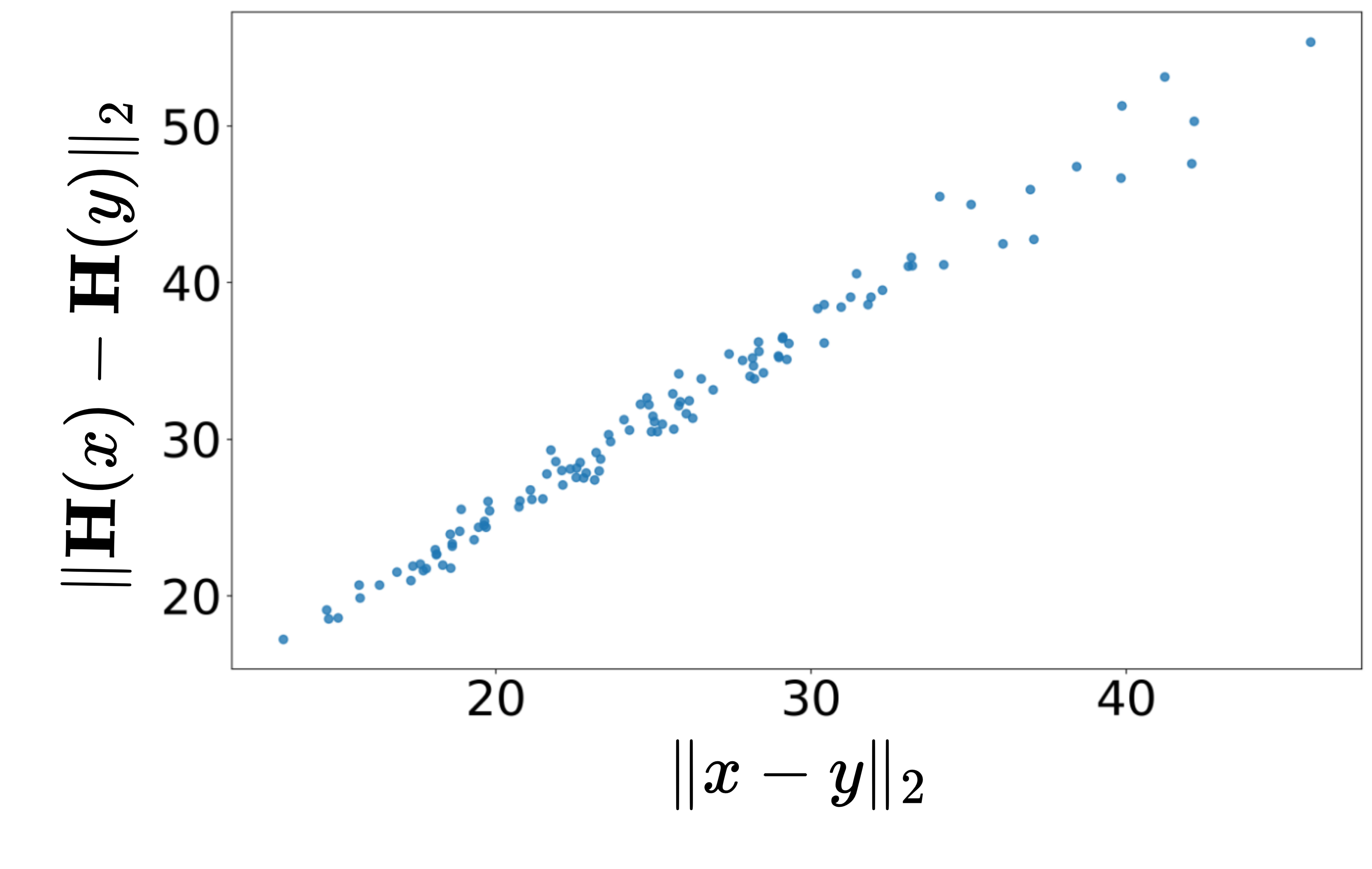}
    \caption{\textit{ETTh1}}
    \label{fig:sub2}
  \end{subfigure}
  \hfill 
  \begin{subfigure}[b]{0.32\textwidth}
    \includegraphics[width=\textwidth]{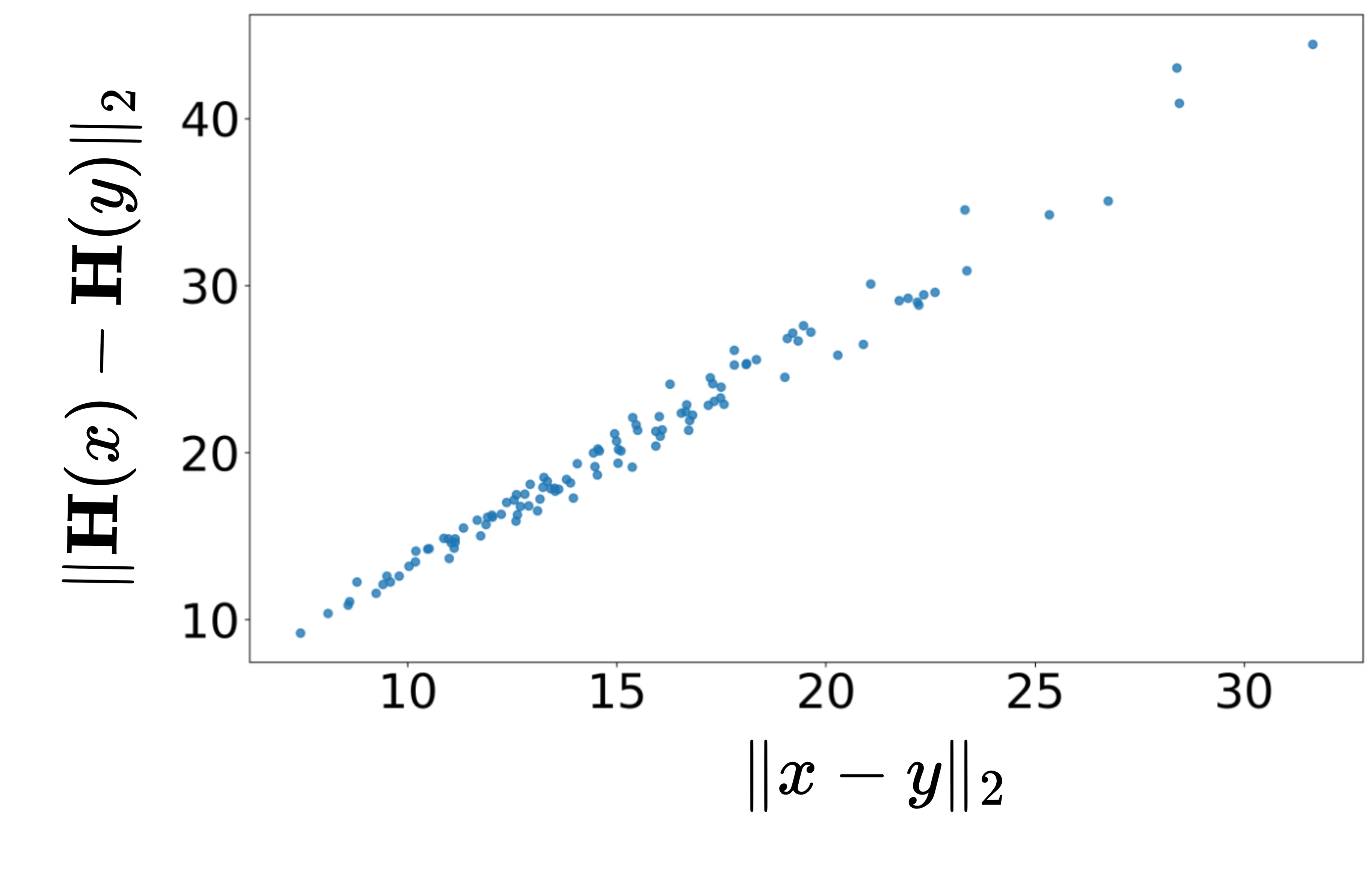}
    \caption{\textit{ETTm1}}
    \label{fig:sub2}
  \end{subfigure}
  \caption{Distance preservation across hyperdimensional mappings of input sequences $\mathbf{H}(x)$ for $\tau=96$}
  \label{dist_pres_96}
\end{figure}

We see a similar observation from Figure \ref{dist_pres_96} regarding distance preservation for long term TSF ($\tau = 96$) on the same datasets. However, we see greater dispersion of points and a greater variation in the linearity of the distance relationship, which may be the cause of the drop in performance for longer horizon forecasting in TSF-HD.

\begin{figure}[ht]
  \centering
  \begin{subfigure}[b]{0.32\textwidth}
    \includegraphics[width=\textwidth]{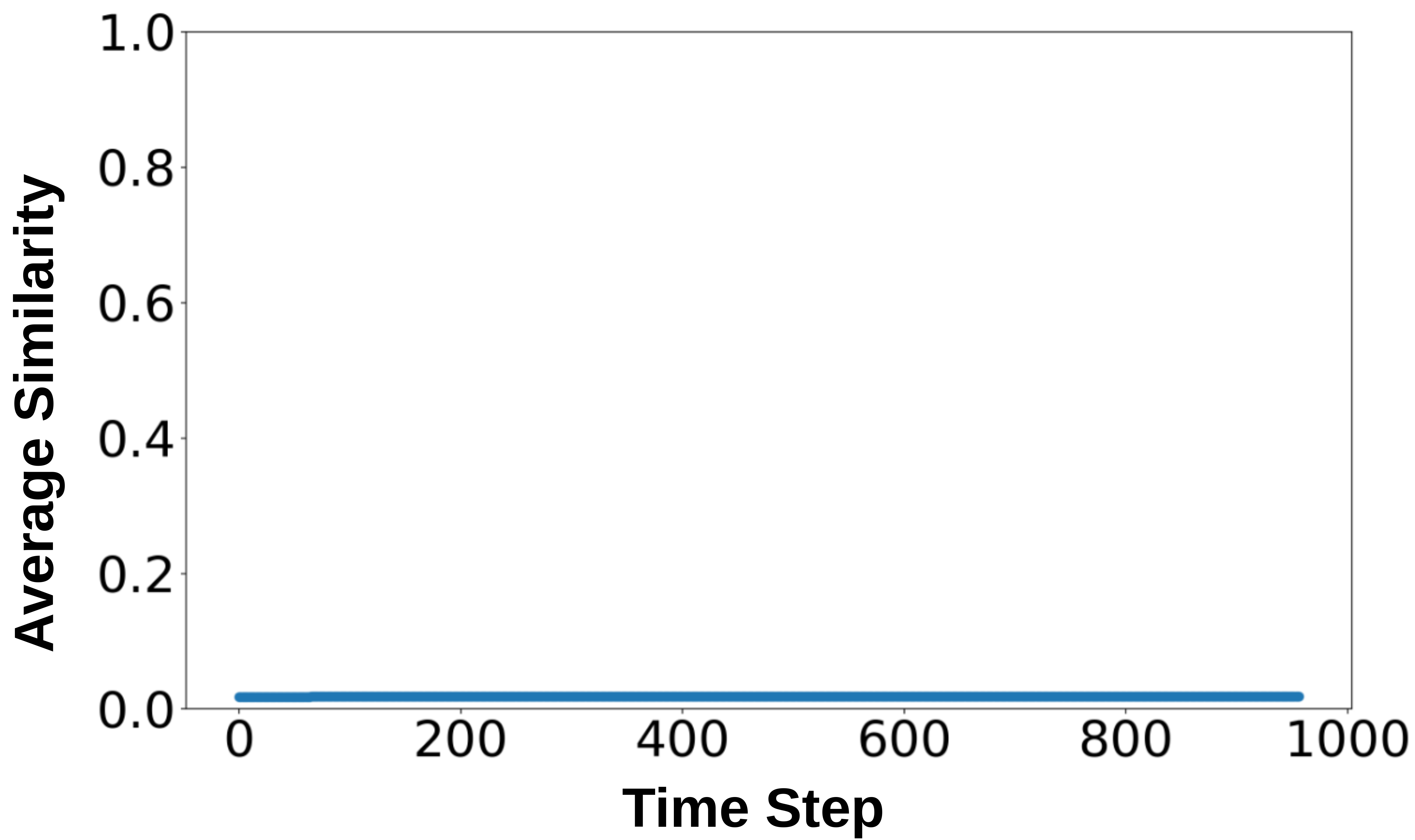}
    \caption{\textit{Exchange}}
    \label{fig:sub1}
  \end{subfigure}
  \hfill 
  \begin{subfigure}[b]{0.32\textwidth}
    \includegraphics[width=\textwidth]{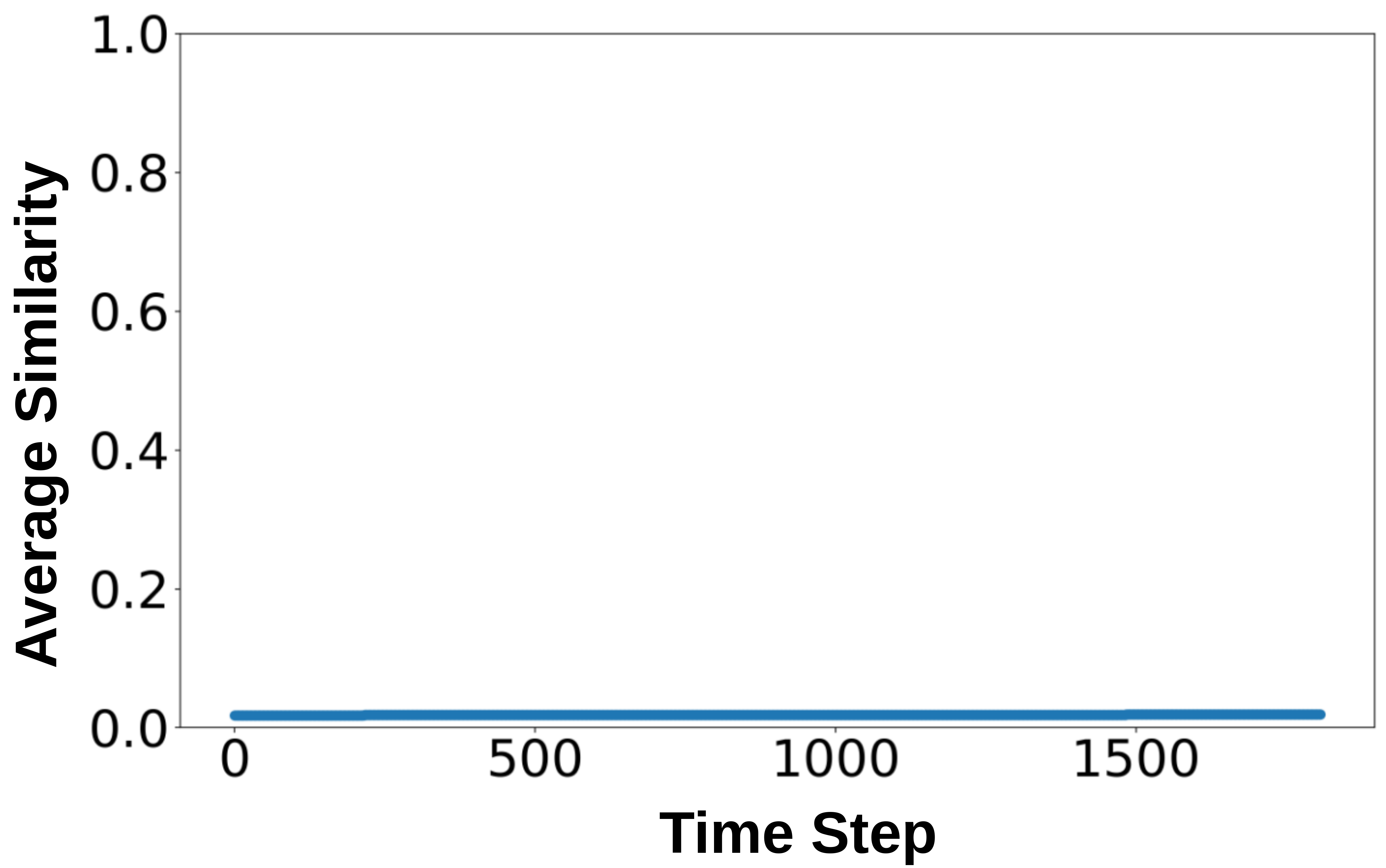}
    \caption{\textit{ETTh1}}
    \label{fig:sub2}
  \end{subfigure}
  \hfill 
  \begin{subfigure}[b]{0.32\textwidth}
    \includegraphics[width=\textwidth]{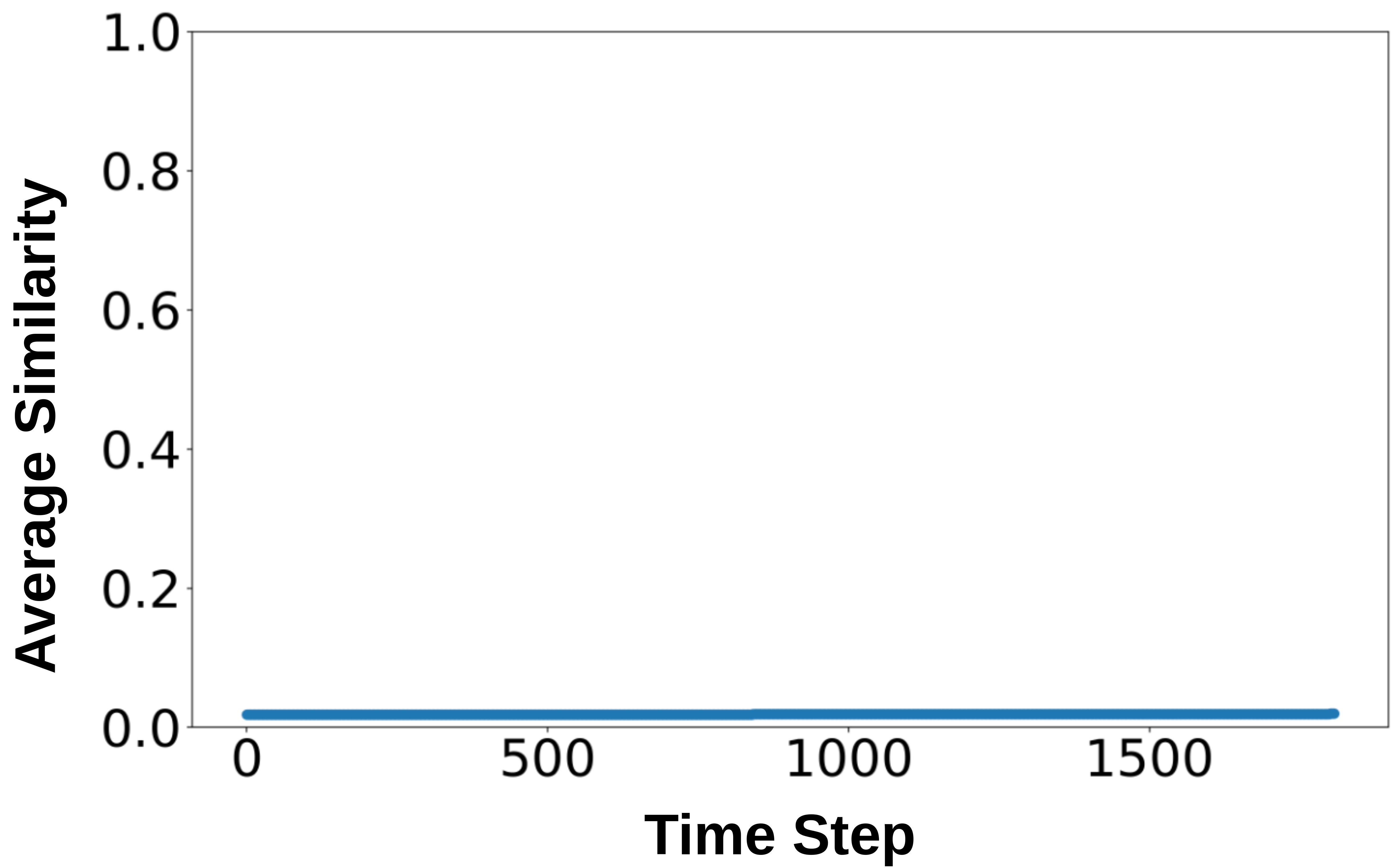}
    \caption{\textit{ETTm1}}
    \label{fig:sub2}
  \end{subfigure}  
  \caption{The mean cosine similarity between rows of the
encoder matrix $W_e$ of $\mathbf{H}(x)$ used to map low-dimensional input
sequences to the hyperdimensional space for $\tau=6$}
  \label{ortho_6}
\end{figure}

Figure \ref{ortho_6} shows the low average cosine similarity ($\leq 0.03$) between hypervectors of the encoder matrix $W_e$, illustrating the orthogonality between $\mathbf{H}$ matrix components for short term forecasting horizon, $\tau = 6$. This illustrates the fact that each component of the encoder is mapping to a different component of the hyperspace, taking advantage of the high dimensionality of the hyperspace compared to the input data.

\begin{figure}[ht]
  \centering
  \begin{subfigure}[b]{0.32\textwidth}
    \includegraphics[width=\textwidth]{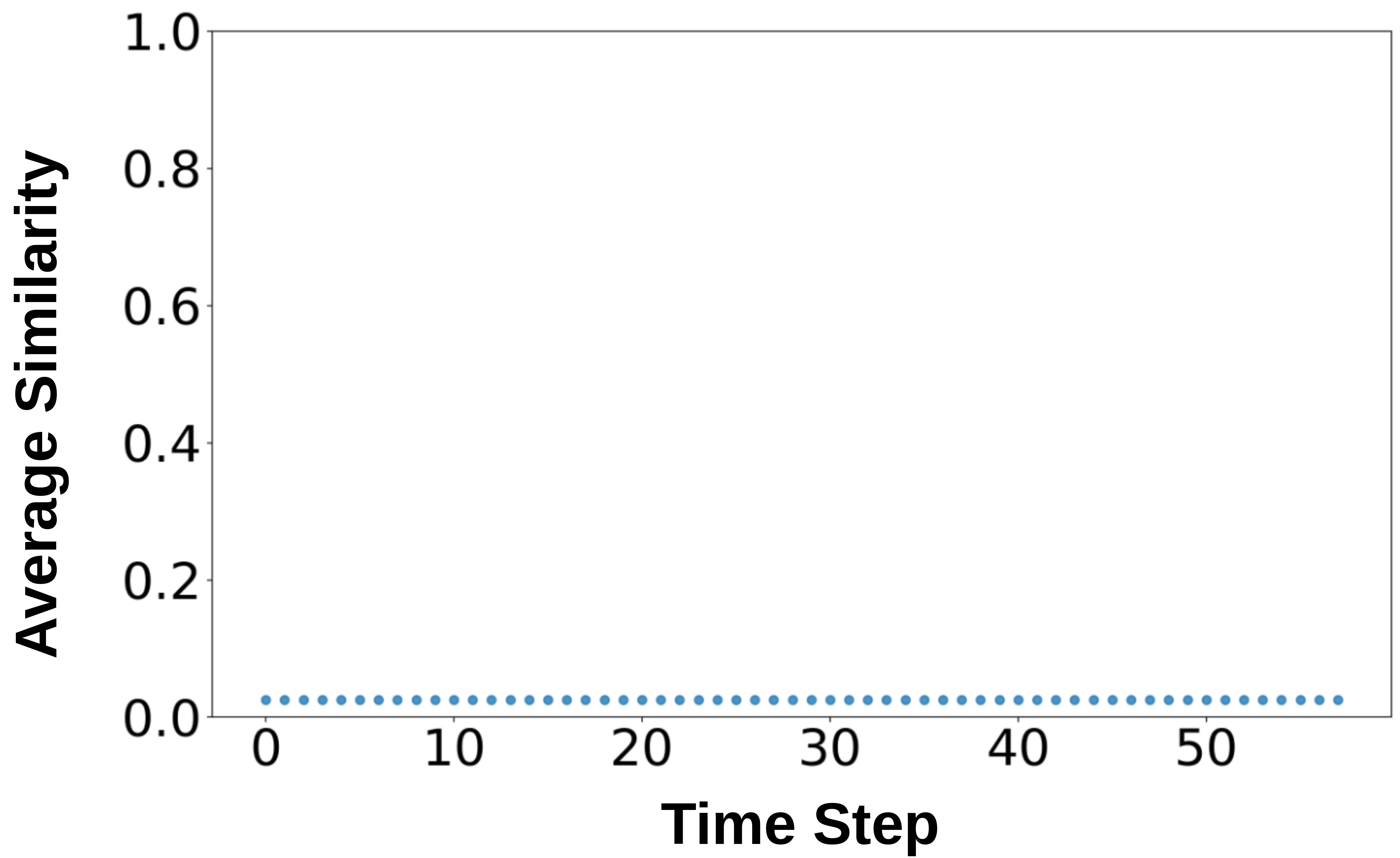}
    \caption{\textit{Exchange}}
    \label{fig:sub1}
  \end{subfigure}
  \hfill 
  \begin{subfigure}[b]{0.32\textwidth}
    \includegraphics[width=\textwidth]{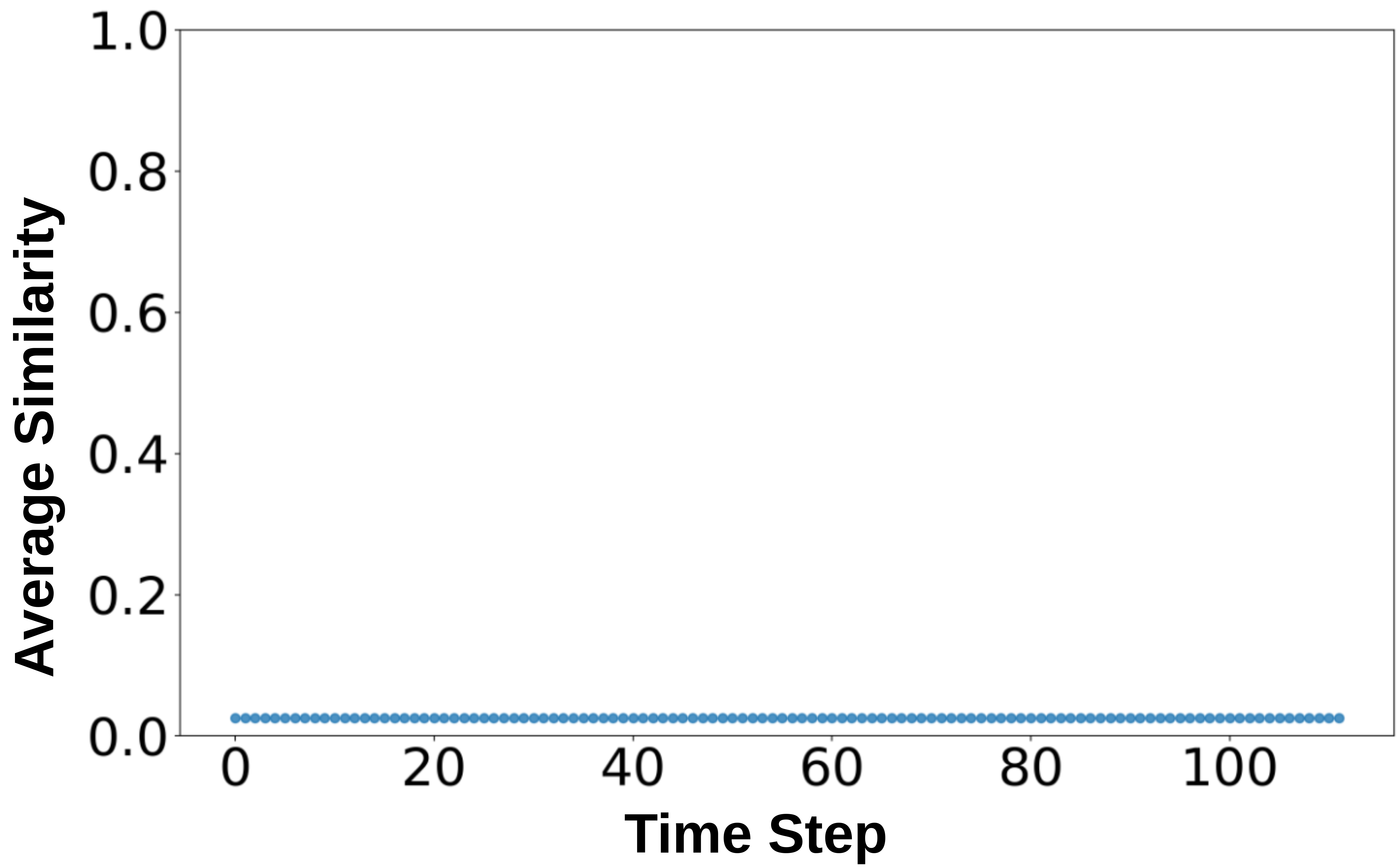}
    \caption{\textit{ETTh1}}
    \label{fig:sub2}
  \end{subfigure}
    \hfill 
  \begin{subfigure}[b]{0.32\textwidth}
    \includegraphics[width=\textwidth]{images/ETTm1_orth.png}
    \caption{\textit{ETTm1}}
    \label{fig:sub2}
  \end{subfigure}
  \caption{The mean cosine similarity between rows of the
encoder matrix $W_e$ of $\mathbf{H}(x)$ used to map low-dimensional input
sequences to the hyperdimensional space for $\tau=96$}
  \label{ortho_96}
\end{figure}

Similar results are observed for hypervector orthogonality in long term TSF $\tau=96$ (see Figure \ref{ortho_96}). However, a similar trend to that between Figures \ref{dist_pres_6} and \ref{dist_pres_96} is also seen, with a higher value of cosine similarity (while still low, at less than 0.05), indicating less efficient use of data thanks to the higher dimensionality of the inputs (larger lookback window). 

\section{Datasets and Evaluation Metrics}
\subsection{Dataset Details}\label{dataset_details}
\begin{table*}[th!]
\begin{center}
\begin{adjustbox}{center, width=0.8\textwidth}
\begin{tabular}{c|cccccc}
\hline
Datasets    & ETTh1 \& ETTh2 & ETTm1 \& ETTm2 & ECL    & WTH    & Exchange& ILI   \\ \hline
Variates    & 7              & 7              & 321    & 12     & 8             & 7     \\
Timesteps   & 17,420         & 69,680         & 26,304 & 52,696 & 7,588         & 966   \\
Granularity & 1hour          & 5min           & 1hour  & 10min  & 1day          & 1week \\ \hline
\end{tabular}
\end{adjustbox}
\caption{The overall information of the 8 popular TSF Datasets}
\label{datasets}
\end{center}
\end{table*}
Details of the real-world benchmark datasets we used are below, with number of variables (dimension of each sample vector), number of time-steps and granularity of sampling in Table \ref{datasets}:
\begin{itemize}
    \item \textbf{ETT}\cite{zhou2021informer}\footnote{\url{https://github.com/zhouhaoyi/ETDataset}} The dataset is composed of two parts: ETTh, which includes hourly data, and ETTm, which features data at 15-minute intervals. Each dataset provides information on seven different features related to oil and load in electricity transformers. This data spans a period from July 2016 to July 2018.  
    \item \textbf{WTH}\footnote{\url{https://www.bgc-jena.mpg.de/wetter/}}  The dataset encompasses 21 weather-related metrics, including air temperature and humidity, meticulously recorded every 10 minutes throughout the year 2020 in Germany.
    \item \textbf{Exchange}\footnote{\url{https://github.com/laiguokun/multivariate-time-series-data}} collects the daily exchange rates of 8 countries from 1990 to 2016.
    \item\textbf{ILI}\footnote{\url{https://gis.cdc.gov/grasp/fluview/fluportaldashboard.html}} his dataset details the proportion of patients presenting with influenza-like illness compared to the total number of patients seen. It comprises weekly records from the U.S. Centers for Disease Control and Prevention, spanning from 2002 to 2021
    \item\textbf{ECL}\footnote{\url{https://archive.ics.uci.edu/ml/datasets/ElectricityLoadDiagrams20112014}} collects the hourly electricity consumption of $321$ clients from $2012$ to $2014$.
\end{itemize}

\begin{figure*}[h]
    \centering
\includegraphics[width=0.7\textwidth]{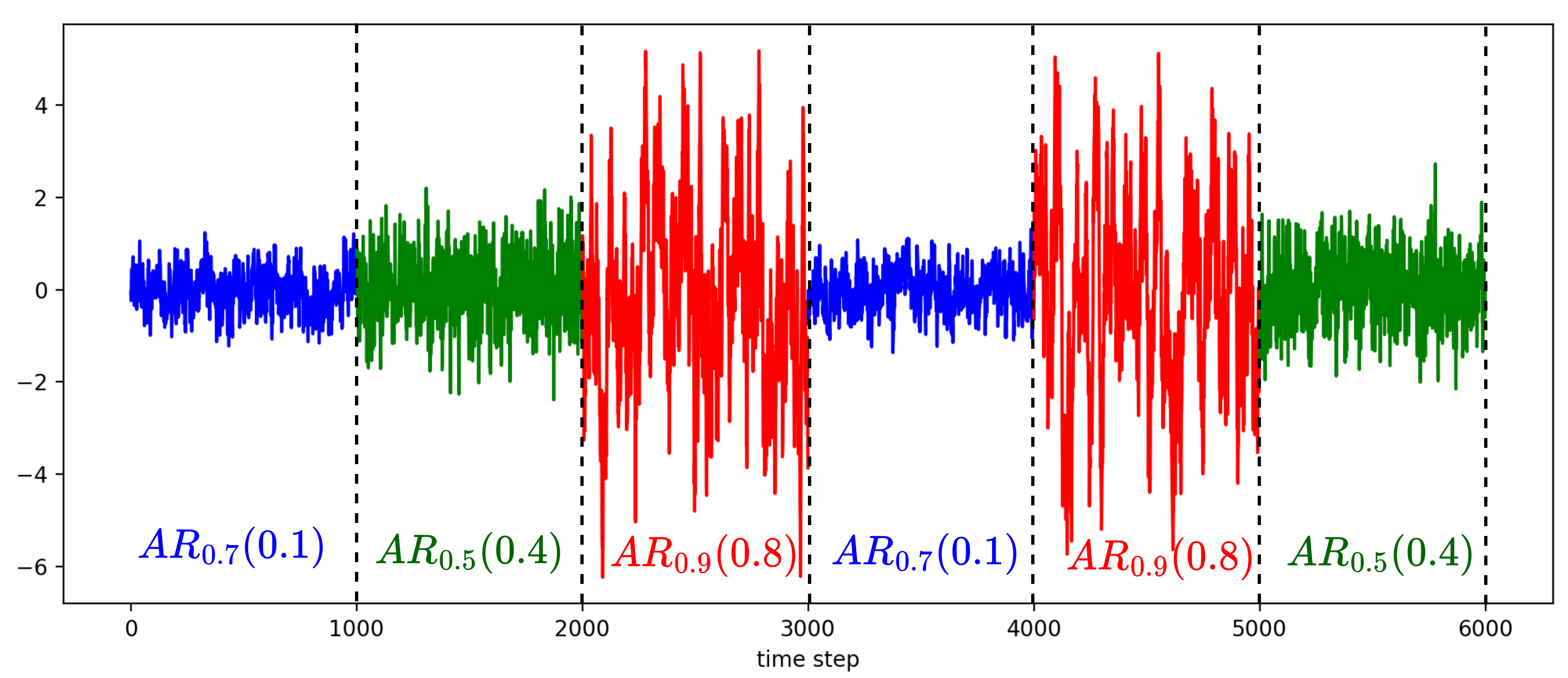}
    \caption{Overview of Synthetic Abrupt time series (S-A)}
    \label{fig:S-A}
\end{figure*}

We also use the Synthetic-Abrupt univariate time series \textbf{S-A} dataset, drawing from \cite{pham2022learning}. It is a concatenation of first-order auto-regressive processes $AR_{\varphi}(\sigma)$ defined as
\begin{equation}
    X_t = \varphi X_{t-1} + \epsilon_t,
\end{equation}
where $\epsilon_t$ is zero-mean Gaussian noise with variance $\sigma$. The first term $X_{0}$ is randomly generated from a zero-mean Gaussian distribution. The S-Abrupt dataset is generated with different processes acting at different time intervals, as described by the following equation:
\begin{equation}
    X_t=
    \begin{cases}
      AR1 = AR_{0.7}(0.1) & \mathrm{if}\ 1 < t \leq 1000 \\
      AR3 = AR_{0.5}(0.4) & \mathrm{if}\ 1000 < t \leq 1999 \\
      AR2 = AR_{0.9}(0.8) & \mathrm{if}\ 2000 < t \leq 2999 \\
      AR1 = AR_{0.7}(0.1) & \mathrm{if}\ 3000 < t \leq 3999 \\
      AR2 = AR_{0.9}(0.8) & \mathrm{if}\ 4000 < t \leq 4999 \\
      AR3 = AR_{0.5}(0.4) & \mathrm{if}\ 5000 < t \leq 5999. \\
    \end{cases}
  \end{equation}
The \textbf{S-A} implementation in this paper is different from the one proposed in \cite{pham2022learning}, as we have changed the variance of $\epsilon_t$ in each of the AR processes to accentuate the severity of the task shifts (concept drift) in the time series. Earlier work \cite{pham2022learning} retained identical noise variance for all AR processes.

Figure \ref{fig:S-A} illustrates the S-A dataset over time with a color associated to each AR process. It begins with a warmup process $AR1$ followed by $AR2$, $AR1$ again, $AR2$ again and $AR3$, to evaluate forecaster adaptation to task shifts \textit{and} recurrence of old tasks. Figure \ref{viz} of Section \ref{adaptation} illustrates concept drift adaptation for TSF-HD (main body of the paper) across this dataset. Section \ref{adaptation} skips the performance analysis for the first appearance of $AR3$ for brevity, but we run the online learning systems across the entire dataset shown in Figure \ref{fig:S-A}. We allow online learning to occur for the first 200 of the 1000 timesteps of each process, followed by comparing predictions and learning online to state of the art for the next 20 time steps (following the setup of \cite{pham2022learning}).

\subsection{Metrics}\label{metrics_appendix}
The metrics used to evaluate forecaster accuracy are $CORR$ (Empirical Correlation  Coefficient) \cite{lai2018modeling} and $RSE$ (Relative Root Squared Error) \cite{lai2018modeling}
detailed here:
\begin{equation}
    RSE(\tilde{X},X) = \frac{\sqrt{\sum_{i=0}^{\tau} (\tilde{x}_i-x_i)^2}}{\sqrt{\sum_{i=0}^{\tau}(x_i-\Bar{X})^2}} 
\end{equation}
\begin{equation}
    CORR(\tilde{X},X) = \frac{1}{d} \sum_{j=0}^d \frac{\sum_{i=0}^{\tau} (x_{i,j}-\Bar{X_j})(\tilde{x}_{i,j}-\Bar{\tilde{X_j}})}{\sum_{i=0}^{\tau} (x_{i,j}-\Bar{X_j})^2 (\tilde{x}_{i,j}-\Bar{\tilde{X_j}})^2}
\end{equation}
in the $RSE$ and $CORR$ equations, $\tilde{x}$ and $x$ refer respectively to the predicted sample and the ground truth sample. $\Bar{\tilde{X}}$ refers to the mean of the entire predicted sequence while $\Bar{X}$ refers to the mean of the entire ground truth sequence. For RSE, a lower value indicates a lower average squared error when compared to the naive forecaster (forecasting all values as the sequence mean), and thus indicates better accuracy. For CORR, the correlation coefficient ranges from 0 to 1 and indicates how well-correlated the forecaster predictions are with the data stream - a higher value indicates more correlated predictions (better precision). 

\section{Baselines}\label{baselines_appendix}

\begin{table*}[ht]
\begin{center}
\begin{adjustbox}{center, width=0.75\textwidth}
\begin{tabular}{cc|cccccccccccc|}
\cline{3-14}
 &
   &
  \multicolumn{2}{c}{{\color[HTML]{5B277D} \textbf{Seq2Seq-HDC}}} &
  \multicolumn{2}{c}{{\color[HTML]{5B277D} \textbf{AR-HDC}}} &
  \multicolumn{2}{c}{{\color[HTML]{2A6099} \textbf{SCINet}}} &
  \multicolumn{2}{c}{{\color[HTML]{2A6099} \textbf{Nlinear}}} &
  \multicolumn{2}{c}{{\color[HTML]{2A6099} \textbf{Naive}}} &
  \multicolumn{2}{c|}{{\color[HTML]{2A6099} \textbf{GBRT}}} \\ \cline{2-14} 
\multicolumn{1}{c|}{} &
  \textit{$\tau$} &
  \textit{\textbf{RSE}} &
  \textit{\textbf{CORR}} &
  \textit{\textbf{RSE}} &
  \textit{\textbf{CORR}} &
  \textit{\textbf{RSE}} &
  \textit{\textbf{CORR}} &
  \textit{\textbf{RSE}} &
  \textit{\textbf{CORR}} &
  \textit{\textbf{RSE}} &
  \textit{\textbf{CORR}} &
  \textit{\textbf{RSE}} &
  \textit{\textbf{CORR}} \\ \hline
\multicolumn{1}{|c|}{} &
  3 &
  0.032 &
  0.999 &
  0.033 &
  0.999 &
  0.053 &
  0.999 &
  {\color[HTML]{3465A4} {\underline{0.017}}} &
  {\color[HTML]{3465A4} {\underline{0.999}}} &
  \textbf{0.011} &
  \textbf{0.999} &
  0.018 &
  0.999 \\
\multicolumn{1}{|c|}{} &
  6 &
  0.043 &
  0.999 &
  0.034 &
  0.999 &
  0.052 &
  0.999 &
  {\color[HTML]{3465A4} {\underline{0.021}}} &
  {\color[HTML]{3465A4} {\underline{0.999}}} &
  \textbf{0.017} &
  \textbf{0.999} &
  0.021 &
  0.999 \\
\multicolumn{1}{|c|}{\multirow{-3}{*}{\textbf{Exchange}}} &
  12 &
  0.053 &
  0.998 &
  0.035 &
  0.999 &
  0.05 &
  0.999 &
  {\color[HTML]{3465A4} {\underline{0.028}}} &
  {\color[HTML]{3465A4} {\underline{0.999}}} &
  \textbf{0.023} &
  \textbf{0.999} &
  0.027 &
  0.999 \\ \hline
\multicolumn{1}{|c|}{} &
  3 &
  0.314 &
  {\color[HTML]{3465A4} {\underline{0.952}}} &
  {\color[HTML]{3465A4} {\underline{0.288}}} &
  \textbf{0.966} &
  \textbf{0.129} &
  0.902 &
  0.331 &
  0.945 &
  0.329 &
  0.945 &
  0.381 &
  0.925 \\
\multicolumn{1}{|c|}{} &
  6 &
  \textbf{0.153} &
  \textbf{0.988} &
  0.305 &
  {\color[HTML]{3465A4} {\underline{0.976}}} &
  {\color[HTML]{3465A4} {\underline{0.154}}} &
  0.875 &
  0.503 &
  0.873 &
  0.493 &
  0.879 &
  0.47 &
  0.882 \\
\multicolumn{1}{|c|}{\multirow{-3}{*}{\textbf{ECL}}} &
  12 &
  \textbf{0.106} &
  \textbf{0.994} &
  {\color[HTML]{3465A4} {\underline{0.121}}} &
  \textbf{0.994} &
  0.144 &
  {\color[HTML]{3465A4} {\underline{0.866}}} &
  0.732 &
  0.731 &
  0.733 &
  0.739 &
  0.596 &
  0.809 \\ \hline
\multicolumn{1}{|c|}{} &
  3 &
  0.142 &
  0.989 &
  {\color[HTML]{3465A4} {\underline{0.128}}} &
  \textbf{0.994} &
  0.179 &
  0.984 &
  \textbf{0.124} &
  {\color[HTML]{3465A4} {\underline{0.992}}} &
  0.125 &
  0.992 &
  0.134 &
  0.991 \\
\multicolumn{1}{|c|}{} &
  6 &
  {\color[HTML]{3465A4} {\underline{0.162}}} &
  0.987 &
  0.177 &
  \textbf{0.991} &
  0.209 &
  0.982 &
  \textbf{0.147} &
  {\color[HTML]{3465A4} {\underline{0.989}}} &
  0.174 &
  0.985 &
  0.167 &
  0.986 \\
\multicolumn{1}{|c|}{\multirow{-3}{*}{\textbf{ETTh2}}} &
  12 &
  0.178 &
  0.984 &
  {\color[HTML]{3465A4} {\underline{0.172}}}&
  \textbf{0.988} &
  0.21 &
  0.978 &
  \textbf{0.169} &
  {\color[HTML]{3465A4} {\underline{0.985}}} &
  0.234 &
  0.973 &
  0.207 &
  0.978 \\ \hline
\multicolumn{1}{|c|}{} &
  3 &
  {\color[HTML]{3465A4} {\underline{0.389}}} &
  \textbf{0.928} &
  \textbf{0.349} &
  {\color[HTML]{3465A4} {\underline{0.92}}} &
  0.406 &
  0.916 &
  0.396 &
  0.917 &
  0.562 &
  0.842 &
  0.623 &
  0.785 \\
\multicolumn{1}{|c|}{} &
  6 &
  0.482 &
  \textbf{0.894} &
  \textbf{0.443} &
  0.875 &
  {\color[HTML]{3465A4} {\underline{0.468}}} &
  {\color[HTML]{3465A4} {\underline{0.885}}} &
  0.47 &
  0.882 &
  0.892 &
  0.609 &
  0.817 &
  0.592 \\
\multicolumn{1}{|c|}{\multirow{-3}{*}{\textbf{ETTh1}}} &
  12 &
  \textbf{0.371} &
  \textbf{0.938} &
  {\color[HTML]{3465A4} {\underline{0.448}}} &
  {\color[HTML]{3465A4} {\underline{0.902}}} &
  0.53 &
  0.86 &
  0.517 &
  0.855 &
  1.227 &
  0.289 &
  1.011 &
  0.362 \\ \hline
\multicolumn{1}{|c|}{} &
  3 &
  {\color[HTML]{3465A4} {\underline{0.111}}}&
  \textbf{0.994} &
  \textbf{0.11} &
  \textbf{0.994} &
  0.302 &
  0.957 &
  0.253 &
  {\color[HTML]{3465A4} {\underline{0.968}}} &
  0.267 &
  0.964 &
  0.422 &
  0.914 \\
\multicolumn{1}{|c|}{} &
  6 &
  \textbf{0.135} &
  \textbf{0.991} &
  {\color[HTML]{3465A4} {\underline{0.148}}} &
  {\color[HTML]{3465A4} {\underline{0.99}}} &
  0.411 &
  0.916 &
  0.338 &
  0.943 &
  0.363 &
  0.934 &
  0.484 &
  0.879 \\
\multicolumn{1}{|c|}{\multirow{-3}{*}{\textbf{ETTm1}}} &
  12 &
  \textbf{0.181} &
  \textbf{0.984} &
  {\color[HTML]{3465A4} {\underline{0.209}}} &
  {\color[HTML]{3465A4} {\underline{0.982}}} &
  0.517 &
  0.86 &
  0.503 &
  0.874 &
  0.538 &
  0.854 &
  0.609 &
  0.793 \\ \hline
\multicolumn{1}{|c|}{} &
  3 &
  0.104 &
  0.994 &
  \textbf{0.086} &
  \textbf{0.998} &
  0.162 &
  0.989 &
  0.081 &
  0.996 &
  {\color[HTML]{3465A4} {\underline{0.087}}} &
  {\color[HTML]{3465A4} {\underline{0.996}}} &
  0.102 &
  0.994 \\
\multicolumn{1}{|c|}{} &
  6 &
  0.136 &
  0.991 &
  {\color[HTML]{3465A4} {\underline{0.114}}} &
  \textbf{0.996} &
  0.134 &
  0.991 &
  0.097 &
  0.995 &
  \textbf{0.103} &
  {\color[HTML]{3465A4} {\underline{0.995}}} &
  0.113 &
  0.993 \\
\multicolumn{1}{|c|}{\multirow{-3}{*}{\textbf{ETTm2}}} &
  12 &
  0.161 &
  0.987 &
  0.171 &
  \textbf{0.993} &
  0.174 &
  0.984 &
  0.121 &
  0.992 &
  \textbf{0.127} &
  {\color[HTML]{3465A4} {\underline{0.992}}} &
  {\color[HTML]{3465A4} {\underline{0.133}}} &
  0.991 \\ \hline
\multicolumn{1}{|c|}{} &
  3 &
  0.671 &
  0.784 &
  0.642 &
  {\color[HTML]{3465A4} {\underline{0.797}}} &
  {\color[HTML]{3465A4} {\underline{0.636}}} &
  0.78 &
  0.724 &
  0.737 &
  0.711 &
  0.742 &
  \textbf{0.596} &
  \textbf{0.803} \\
\multicolumn{1}{|c|}{} &
  6 &
  0.712 &
  {\color[HTML]{3465A4} {\underline{0.773}}} &
  0.674 &
  \textbf{0.785} &
  \textbf{0.647} &
  {\color[HTML]{3465A4} {\underline{0.773}}}&
  0.797&
  0.68 &
  0.788 &
  0.681 &
  {\color[HTML]{3465A4} {\underline{0.65}}} &
  0.763 \\
\multicolumn{1}{|c|}{\multirow{-3}{*}{\textbf{WTH}}} &
  12 &
  0.665 &
  {\color[HTML]{3465A4} {\underline{0.799}}} &
  {\color[HTML]{3465A4} {\underline{0.651}}} &
  \textbf{0.802} &
  \textbf{0.645} &
  0.77 &
  0.854 &
  0.633 &
  0.834 &
  0.635 &
  0.68 &
  0.737 \\ \hline
\multicolumn{1}{|c|}{} &
  3 &
  0.203 &
  0.979 &
  0.135 &
  \textbf{0.998} &
  0.23 &
  0.983 &
  0.118 &
  0.993 &
  {\color[HTML]{3465A4} {\underline{0.103}}} &
  {\color[HTML]{3465A4} {\underline{0.996}}} &
  \textbf{0.093} &
  {\color[HTML]{3465A4} {\underline{0.996}}} \\
\multicolumn{1}{|c|}{} &
  6 &
  0.247 &
  0.969 &
  0.202 &
  \textbf{0.995} &
  0.364 &
  0.948 &
  \textbf{0.153} &
  0.988 &
  0.169 &
  0.99 &
  {\color[HTML]{3465A4} {\underline{0.155}}} &
  {\color[HTML]{3465A4} {\underline{0.991}}} \\
\multicolumn{1}{|c|}{\multirow{-3}{*}{\textbf{ILI}}} &
  12 &
  0.296 &
  0.956 &
  0.297 &
  \textbf{0.989} &
  0.383 &
  0.953 &
  0.196 &
  0.981 &
  {\color[HTML]{3465A4} {\underline{0.252}}} &
  0.982 &
  \textbf{0.236} &
  {\color[HTML]{3465A4} {\underline{0.985}}} \\ \hline
\end{tabular}
\end{adjustbox}
\end{center}
\caption{Short Term Time Series Forecasting Performance of AR-HDC \& Seq2Seq-HDC compared to other offline baselines. We report the mean of RSE and CORR of the experiments. The results in bold are the best and in blue and underlined are second best}
\label{short-term-offline}
\end{table*}
\begin{table*}[ht]
\begin{center}
\begin{adjustbox}{center, width=0.75\textwidth}
\begin{tabular}{cc|cccccccccccc|}
\cline{3-14}
 &
   &
  \multicolumn{2}{c}{{\color[HTML]{5B277D} \textbf{Seq2Seq-HDC}}} &
  \multicolumn{2}{c}{{\color[HTML]{5B277D} \textbf{AR-HDC}}} &
  \multicolumn{2}{c}{{\color[HTML]{2A6099} \textbf{SCINet}}} &
  \multicolumn{2}{c}{{\color[HTML]{2A6099} \textbf{Nlinear}}} &
  \multicolumn{2}{c}{{\color[HTML]{2A6099} \textbf{Naive}}} &
  \multicolumn{2}{c|}{{\color[HTML]{2A6099} \textbf{GBRT}}} \\ \cline{2-14} 
\multicolumn{1}{c|}{} &
  \textit{tau} &
  \textit{\textbf{RSE}} &
  \textit{\textbf{CORR}} &
  \textit{\textbf{RSE}} &
  \textit{\textbf{CORR}} &
  \textit{\textbf{RSE}} &
  \textit{\textbf{CORR}} &
  \textit{\textbf{RSE}} &
  \textit{\textbf{CORR}} &
  \textit{\textbf{RSE}} &
  \textit{\textbf{CORR}} &
  \textit{\textbf{RSE}} &
  \textit{\textbf{CORR}} \\ \hline
\multicolumn{1}{|c|}{} &
  96 &
  {\color[HTML]{3465A4} {\underline{0.601}}} &
  \textbf{0.846} &
  \textbf{0.568} &
  {\color[HTML]{3465A4} {\underline{0.84}}} &
  {\color[HTML]{000000} 0.698} &
  {\color[HTML]{000000} 0.753} &
  0.643 &
  0.7739 &
  1.208 &
  0.273 &
  1.01 &
  0.347 \\
\multicolumn{1}{|c|}{} &
  192 &
  {\color[HTML]{000000} 0.678} &
  {\color[HTML]{3465A4} {\underline{0.803}}} &
  \textbf{0.599} &
  \textbf{0.827} &
  {\color[HTML]{000000} 0.763} &
  {\color[HTML]{000000} 0.7} &
  {\color[HTML]{2A6099} {\underline{0.671}}} &
  0.752 &
  1.226 &
  0.255 &
  1.02 &
  0.328 \\
\multicolumn{1}{|c|}{\multirow{-3}{*}{\textbf{ETTh1}}} &
  384 &
  {\color[HTML]{000000} 0.797} &
  {\color[HTML]{3465A4} {\underline{0.739}}} &
  \textbf{0.628} &
  \textbf{0.831} &
  {\color[HTML]{000000} 0.864} &
  {\color[HTML]{000000} 0.66} &
  {\color[HTML]{2A6099} {\underline{0.687}}} &
  0.735 &
  1.239 &
  0.247 &
  1.03 &
  0.321 \\ \hline
\multicolumn{1}{|c|}{} &
  96 &
  {\color[HTML]{3465A4} {\underline{0.315}}} &
  {\color[HTML]{3465A4} {\underline{0.951}}} &
  {\color[HTML]{000000} \textbf{0.297}} &
  \textbf{0.965} &
  0.585 &
  0.817 &
  0.587 &
  0.813 &
  1.127 &
  0.345 &
  0.985 &
  0.405 \\
\multicolumn{1}{|c|}{} &
  192 &
  {\color[HTML]{3465A4} {\underline{0.392}}} &
  {\color[HTML]{3465A4} {\underline{0.925}}} &
  \textbf{0.373} &
  {\color[HTML]{000000} \textbf{0.949}} &
  0.672 &
  0.767 &
  0.658 &
  0.762 &
  1.143 &
  0.323 &
  1.01 &
  0.375 \\
\multicolumn{1}{|c|}{\multirow{-3}{*}{\textbf{ETTm1}}} &
  384 &
  {\color[HTML]{3465A4} {\underline{0.465}}} &
  {\color[HTML]{3465A4} {\underline{0.893}}} &
  {\color[HTML]{000000} \textbf{0.431}} &
  \textbf{0.924} &
  1.064 &
  0.602 &
  0.731 &
  0.698 &
  1.166 &
  0.296 &
  1.033 &
  0.343 \\ \hline
\multicolumn{1}{|c|}{} &
  24 &
  0.288 &
  0.957 &
  {\color[HTML]{3465A4} {\underline{0.229}}} &
  \textbf{0.986} &
  0.382 &
  {\color[HTML]{000000} 0.934} &
  \textbf{0.173} &
  {\color[HTML]{3465A4} {\underline{0.985}}} &
  0.304 &
  0.98 &
  0.277 &
  0.983 \\
\multicolumn{1}{|c|}{} &
  36 &
  {\color[HTML]{000000} 0.398} &
  0.918 &
  {\color[HTML]{3465A4} {\underline{0.199}}} &
  \textbf{0.99} &
  0.42 &
  {\color[HTML]{000000} 0.969} &
  \textbf{0.176} &
  {\color[HTML]{3465A4} {\underline{0.985}}} &
  0.334 &
  0.981 &
  0.307 &
  0.984 \\
\multicolumn{1}{|c|}{\multirow{-3}{*}{\textbf{ILI}}} &
  48 &
  0.448 &
  0.901 &
  {\color[HTML]{3465A4} {\underline{0.204}}} &
  \textbf{0.989} &
  0.481 &
  {\color[HTML]{000000} 0.965} &
  \textbf{0.185} &
  {\color[HTML]{3465A4} {\underline{0.984}}} &
  0.33 &
  0.968 &
  0.303 &
  0.972 \\ \hline
\multicolumn{1}{|c|}{} &
  96 &
  {\color[HTML]{000000} 0.211} &
  0.978 &
  {\color[HTML]{3465A4} {\underline{0.097}}} &
  {\color[HTML]{3465A4} {\underline{0.997}}} &
  0.092 &
  0.996 &
  \textbf{0.062} &
  \textbf{0.998} &
  0.062 &
  \textbf{0.998} &
  0.062 &
  \textbf{0.998} \\
\multicolumn{1}{|c|}{} &
  192 &
  \textit{0.318} &
  0.948 &
  0.155 &
  {\color[HTML]{3465A4} {\underline{0.994}}} &
  0.159 &
  0.989 &
  {\color[HTML]{3465A4} {\underline{0.089}}} &
  \textbf{0.996} &
  \textbf{0.088} &
  \textbf{0.996} &
  0.088 &
  \textbf{0.996} \\
\multicolumn{1}{|c|}{\multirow{-3}{*}{\textbf{Exchange}}} &
  224 &
  0.328 &
  0.945 &
  {\color[HTML]{000000} 0.137} &
  {\color[HTML]{3465A4} {\underline{0.995}}} &
  0.17 &
  0.987 &
  {\color[HTML]{000000} 0.096} &
  \textbf{0.996} &
  {\color[HTML]{3465A4} {\underline{0.091}}} &
  \textbf{0.996} &
  \textbf{0.09} &
  \textbf{0.996} \\ \hline
\multicolumn{1}{|c|}{} &
  96 &
  \textbf{0.177} &
  {\color[HTML]{3465A4} {\underline{0.985}}} &
  {\color[HTML]{3465A4} {\underline{0.206}}} &
  \textbf{0.989} &
  {\color[HTML]{000000} 0.364} &
  {\color[HTML]{000000} 0.933} &
  0.726 &
  0.736 &
  0.722 &
  0.735 &
  0.571 &
  0.823 \\
\multicolumn{1}{|c|}{} &
  192 &
  \textbf{0.214} &
  {\color[HTML]{3465A4} {\underline{0.977}}} &
  {\color[HTML]{3465A4} {\underline{0.294}}} &
  {\color[HTML]{000000} \textbf{0.979}} &
  0.551 &
  0.857 &
  0.731 &
  0.733 &
  0.721 &
  0.736 &
  0.572 &
  0.822 \\
\multicolumn{1}{|c|}{\multirow{-3}{*}{\textbf{ECL}}} &
  384 &
  \textbf{0.282} &
  {\color[HTML]{3465A4} {\underline{0.962}}} &
  {\color[HTML]{3465A4} {\underline{0.302}}} &
  {\color[HTML]{000000} \textbf{0.979}} &
  0.796 &
  0.631 &
  0.737 &
  0.729 &
  0.729 &
  0.732 &
  0.581 &
  0.817 \\ \hline
\multicolumn{1}{|c|}{} &
  96 &
  0.697 &
  {\color[HTML]{3465A4} {\underline{0.793}}} &
  \textbf{0.633} &
  \textbf{0.796} &
  {\color[HTML]{3465A4} {\underline{0.651}}} &
  0.767 &
  {\color[HTML]{000000} 0.898} &
  {\color[HTML]{000000} 0.593} &
  0.915 &
  0.585 &
  0.746 &
  0.687 \\
\multicolumn{1}{|c|}{} &
  192 &
  {\color[HTML]{3465A4} {\underline{0.723}}} &
  0.784 &
  {\color[HTML]{000000} \textbf{0.632}} &
  {\color[HTML]{3465A4} {\underline{0.791}}} &
  0.691 &
  0.744 &
  0.906 &
  \textbf{0.856} &
  0.925 &
  0.575 &
  0.756 &
  0.678 \\
\multicolumn{1}{|c|}{\multirow{-3}{*}{\textbf{WTH}}} &
  384 &
  0.777 &
  0.762 &
  {\color[HTML]{3465A4} {\underline{0.653}}} &
  {\color[HTML]{3465A4} {\underline{0.776}}} &
  0.777 &
  0.692 &
  \textbf{0.579} &
  {\color[HTML]{000000} \textbf{0.916}} &
  {\color[HTML]{000000} 0.938} &
  0.566 &
  0.767 &
  0.669 \\ \hline
\end{tabular}
\end{adjustbox}
\end{center}
\caption{Long Term Time Series Forecasting Performance of AR-HDC \& Seq2Seq-HDC compared to the offline baseline. We report the mean of RSE and CORR of the experiments. The results in bold are the best and in blue and underlined are second best}
\label{long-term-offline}
\end{table*}

The details of the baselines we have evaluated TSF-HD against are shown below:
\begin{itemize}
    \item \textbf{OnlineTCN}\footnote{\url{https://github.com/salesforce/fsnet}\label{note1}} uses a standard TCN backbone \cite{woo2022cost} with 10 hidden layers, each of which has two stacks of residual convolution filters.
    \item $\textbf{ER}^{\ref{note1}}$ \cite{chaudhry2019tiny} ER enhances OnlineTCN by adding episodic memory to mix old and new learning samples.
    \item $\textbf{DER++}^{\ref{note1}}$\cite{buzzega2020dark} augments the standard ER with a $\ell_2$ knowledge distillation loss on the previous logits.
    \item $\textbf{FSNet}^{\ref{note1}}$ \cite{pham2022learning} or Fast and Slow learning Network is an online learning technique combining rapid adaptation to new data and memory recall of past events. 
    \item \textbf{Informer}\footnote{\url{https://github.com/zhouhaoyi/Informer2020}} \cite{zhou2021informer} is a transformer-based model employing self-attention for efficiency and a generative decoder for accurate predictions.
    \item \textbf{SCINet}\footnote{\url{https://github.com/cure-lab/SCINet}}\cite{liu2022scinet} is an architecture that enhances TSF by recursively downsampling and convolving data to extract complex temporal features.
    \item \textbf{NLinear}\footnote{\url{https://github.com/cure-lab/LTSF-Linear}\label{note2}} It enhances LTSF-Linear's \cite{zeng2023transformers} performance on shifting datasets by normalizing inputs through subtraction and addition around a linear layer.
    \item $\textbf{Naive}^{\ref{note2}}$ is a naive direct multi-step  which repeats the last value in the
    look-back window.
    \item $\textbf{GBRT}^{\ref{note2}}$ is the classical Gradient Boosting Regression Trees algorithm \cite{friedman2001greedy}.
\end{itemize}
The online learning and transformer baselines' evaluation against TSFHD are shown in Section \ref{precision}, while the results of the offline learning and classical algorithms' comparison to TSF-HD are shown in Appendix \ref{power_results}.

\section{Additional results}\label{power_results}

In Table \ref{short-term-offline} we compare Seq2Seq-HDC and AR-HDC to convolutional, linear, naive and gradient boosting approaches for short term TSF. Aside from the Exchange Rate dataset, where the naive and the linear models slightly outperform our techniques (a similar observation is seen in \cite{zeng2023transformers}), TSF-HD (either AR-HDC or Seq2Seq-HDC) outperforms all offline methods in either the CORR or RSE metrics.

In Table \ref{long-term-offline}, AR-HDC outperforms all offline baselines across most test cases, once again with the exception of the Exchange dataset. In this specific case, the Linear model (NLinear \cite{zeng2023transformers}) and the naive approach prove to be more precise than other techniques. This outcome aligns with previous results where the same baseline surpassed our models in short-term TSF, and aligns with prior work. 

For the majority of test cases, AR-HDC demonstrates superior performance over Seq2Seq-HDC. Notably, AR-HDC, having fewer parameters, tends to converge more rapidly than Seq2Seq-HDC, particularly for long-term TSF. This faster convergence is likely due to underfitting, especially when there are fewer learning steps (i.e., the number of times the samples are presented) in long-term TSF, as the time series samples should not overlap and TSF-HD is trained periodically on the samples it predicted up to the forecast horizon - the larger the horizon, the less frequently TSF-HD models are trained. 

In contrast to the linear model (NLinear) which projects the $T$-dimensional input space into a $\tau$-dimensional prediction space, AR-HDC and Seq2Seq-HDC first project the $T$-dimensional space into an hyperspace of D-dimension ($D\gg max(T,\tau))$ then project it back to the $\tau$-dimensional space. This operation extracts further information from the time series and better retains information about old tasks, allowing better precision for almost all the short term and long term TSF cases save for Exchange.

\begin{table*}[h]
\begin{center}
\begin{adjustbox}{width=0.75\textwidth}
\begin{tabular}{cc|cccccc|}
\cline{3-8}
\multicolumn{2}{c|}{} &
  {\color[HTML]{5B277D} \textbf{AR-HDC}} &
  {\color[HTML]{5B277D} \textbf{Seq2Seq-HDC}} &
  {\color[HTML]{ACB20C} \textbf{FSNet}} &
  {\color[HTML]{ACB20C} \textbf{ER}} &
  {\color[HTML]{ACB20C} \textbf{DER++}} &
  {\color[HTML]{ACB20C} \textbf{OnlineTCN}} \\ \hline
\multicolumn{1}{|c|}{} &
  3 &
  {\color[HTML]{3465A4 } {$\underline{0.073}^{\underline{\pm 0.018}}$}} &
  $\textbf{0.037}^{\pm \textbf{0.021}}$ &
  $0.356^{\pm 0.099}$ &
  $0.364^{\pm 0.031}$ &
  $0.365^{\pm 0.036}$ &
  $0.169^{\pm 0.079}$ \\
\multicolumn{1}{|c|}{\multirow{-2}{*}{\textbf{ECL}}} &
  96 &
  $5.233^{\pm 0.669}$ &
  $\textbf{0.0828}^{\pm \textbf{0.041}}$ &
  $1.44^{\pm 0.277}$ &
  $1.791^{\pm 0.061}$ &
  $1.791^{\pm 0.065}$ &
  {\color[HTML]{3465A4 } {$\underline{0.991}^{\underline{\pm 0.097}}$}} \\ \hline
\multicolumn{1}{|c|}{} &
  3 &
  $\textbf{0.0144}^{\pm \textbf{0.001}}$ &
  {\color[HTML]{3465A4 } {$\underline{0.065}^{\underline{\pm 0.011}}$}} &
  $0.47^{\pm 0.122}$ &
  $0.33^{\pm 0.024}$ &
  $0.335^{\pm 0.023}$ &
  $0.212^{\pm 0.078}$ \\
\multicolumn{1}{|c|}{\multirow{-2}{*}{\textbf{ETTh1}}} &
  96 &
  $0.867^{\pm 0.03}$ &
  $\textbf{0.071}^{\pm \textbf{0.038}}$ &
  $0.530^{\pm 0.209}$ &
  $0.696^{\pm 0.375}$ &
  $0.697^{\pm 0.399}$ &
  {\color[HTML]{3465A4 } {$\underline{0.295}^{\underline{\pm 0.125}}$}} \\ \hline
\multicolumn{1}{|c|}{} &
  3 &
  $\textbf{0.016}^{\pm \textbf{0.002}}$ &
  {\color[HTML]{3465A4 } {$\underline{0.066}^{\underline{\pm 0.013}}$}} &
  $0.452^{\pm 0.121}$ &
  $0.334^{\pm 0.024}$ &
  $0.337^{\pm 0.025}$ &
  $0.213^{\pm 0.079}$ \\
\multicolumn{1}{|c|}{\multirow{-2}{*}{\textbf{Exchange}}} &
  96 &
  $0.937^{\pm 0.347}$ &
  $\textbf{0.073}^{\pm \textbf{0.041}}$ &
  $0.538^{\pm 0.227}$ &
  $0.702^{\pm 0.036}$ &
  $0.701^{\pm 0.039}$ &
  {\color[HTML]{3465A4 } {$\underline{0.273}^{\underline{\pm 0.093}}$}} \\ \hline
\multicolumn{1}{|c|}{} &
  3 &
  $\textbf{0.015}^{\pm \textbf{0.003}}$ &
  {\color[HTML]{3465A4 } {$\underline{0.069}^{\underline{\pm 0.004}}$}} &
  $0.329^{\pm 0.09}$&
  $0.304^{\pm 0.035}$ &
  $0.305^{\pm 0.308}$ &
  $0.149^{\pm 0.082}$ \\
\multicolumn{1}{|c|}{\multirow{-2}{*}{\textbf{WTH}}} &
  96 &
  $1.171^{\pm 0.654}$ &
  $\textbf{0.082}^{\pm \textbf{0.038}}$ &
  $0.678^{\pm 0.316}$ &
  $0.730^{\pm 0.043}$ &
  $0.723^{\pm 0.046}$ &
  {\color[HTML]{3465A4 } {$\underline{0.305}^{\underline{\pm 0.117}}$}} \\ \hline
\end{tabular}
\end{adjustbox}
\caption{\small Latency of Online learning models on RaspberryPI ($mean^{\pm std}$)}
\label{latency-cpu}
\end{center}
\end{table*}
\begin{table*}[h]
\begin{center}
\begin{adjustbox}{width=0.75\textwidth}
\begin{tabular}{cc|cccccc|}
\cline{3-8}
\multicolumn{2}{c|}{} &
  {\color[HTML]{5B277D} \textbf{AR-HDC}} &
  {\color[HTML]{5B277D} \textbf{Seq2Seq-HDC}} &
  {\color[HTML]{ACB20C} \textbf{FSNet}} &
  {\color[HTML]{ACB20C} \textbf{ER}} &
  {\color[HTML]{ACB20C} \textbf{DER++}} &
  {\color[HTML]{ACB20C} \textbf{OnlineTCN}} \\ \hline
\multicolumn{1}{|c|}{} &
  3 &
  $4.715^{\pm 0.976}$ &
  $\textbf{4.153}^{\pm \textbf{0.932}}$ &
  $4.827^{\pm 0.742}$ &
  $4.881^{\pm 0.746}$ &
  $4.914^{\pm 0.739}$ &
  {\color[HTML]{3465A4 } {$\underline{4.471}^{\underline{\pm 0.817}}$}} \\
\multicolumn{1}{|c|}{\multirow{-2}{*}{\textbf{ECL}}} &
  96 &
  $5.548^{\pm 0.506}$ &
  $\textbf{4.762}^{\pm \textbf{0.816}}$ &
  $4.923^{\pm 0.582}$ &
  $5.07^{\pm 0.646}$ &
  $5.077^{\pm 0.649}$ &
  {\color[HTML]{3465A4 } {$\underline{4.914}^{\underline{\pm 0.552}}$}} \\ \hline
\multicolumn{1}{|c|}{} &
  3 &
  $\textbf{3.508}^{\pm \textbf{0.486}}$ &
  {\color[HTML]{3465A4 } {$\underline{4.236}^{\underline{\pm 0.904}}$}} &
  $4.766^{\pm 0.637}$ &
  $4.833^{\pm 0.795}$ &
  $4.906^{\pm 0.788}$ &
  $4.567^{\pm 0.783}$ \\
\multicolumn{1}{|c|}{\multirow{-2}{*}{\textbf{ETTh1}}} &
  96 &
  $5.286^{\pm 0.203}$ &
  $\textbf{4.419}^{\pm \textbf{0.967}}$ &
  $4.961^{\pm 0.591}$ &
  $5.194^{\pm 0.753}$ &
  $5.247^{\pm 0.743}$ &
  {\color[HTML]{3465A4 } {$\underline{4.858}^{\underline{\pm 0.698}}$}} \\ \hline
\multicolumn{1}{|c|}{} &
  3 &
  $\textbf{3.395}^{\pm \textbf{0.534}}$ &
  {\color[HTML]{3465A4 } {$\underline{4.246}^{\underline{\pm 0.945}}$}} &
  $4.855^{\pm 0.635}$ &
  $4.974^{\pm 0.761}$ &
  $4.982^{\pm 0.732}$ &
  $4.654^{\pm 0.748}$ \\
\multicolumn{1}{|c|}{\multirow{-2}{*}{\textbf{Exchange}}} &
  96 &
  $5.257^{\pm 0.511}$ &
  $\textbf{4.437}^{\pm \textbf{0.855}}$ &
  $4.975^{\pm 0.642}$ &
  $5.244^{\pm 0.732}$ &
  $5.259^{\pm 0.742}$ &
  {\color[HTML]{3465A4 } {$\underline{4.788}^{\underline{\pm 0.804}}$}} \\ \hline
\multicolumn{1}{|c|}{} &
  3 &
  $\textbf{3.457}^{\pm \textbf{0.469}}$ &
  {\color[HTML]{3465A4 } {$\underline{3.971}^{\underline{\pm 1.122}}$}} &
  $4.752^{\pm 0.788}$ &
  $4.903^{\pm 0.791}$ &
  $4.804^{\pm 0.901}$ &
  $4.338^{\pm 0.919}$ \\
\multicolumn{1}{|c|}{\multirow{-2}{*}{\textbf{WTH}}} &
  96 &
  $5.272^{\pm 0.429}$ &
  $\textbf{4.450}^{\pm \textbf{0.87}}$ &
  $4.931^{\pm 0.679}$ &
  $5.255^{\pm 0.716}$ &
  $5.208^{\pm 0.768}$ &
  {\color[HTML]{3465A4 } {$\underline{4.818}^{\underline{\pm 0.725}}$}} \\ \hline
\end{tabular}
\end{adjustbox}
\caption{\small Power consumption of Online learning models on RaspberryPI ($mean^{\pm std}$)}
\label{power-cpu}
\end{center}
\end{table*}
\begin{table*}[h]
\begin{center}
\begin{adjustbox}{width=0.75\textwidth}
\begin{tabular}{cc|cccccc|}
\cline{3-8}
\multicolumn{2}{c|}{} &
  {\color[HTML]{5B277D} \textbf{AR-HDC}} &
  {\color[HTML]{5B277D} \textbf{Seq2Seq-HDC}} &
  {\color[HTML]{ACB20C} \textbf{FSNet}} &
  {\color[HTML]{ACB20C} \textbf{ER}} &
  {\color[HTML]{ACB20C} \textbf{DER++}} &
  {\color[HTML]{ACB20C} \textbf{OnlineTCN}} \\ \hline
\multicolumn{1}{|c|}{} &
  3 &
  {\color[HTML]{3465A4 } {$\underline{0.035}^{\underline{\pm0.008}}$}} &
  $\textbf{0.018}^{\pm\textbf{0.008}}$ &
  $0.637^{\pm1.638}$ &
  $0.229^{\pm0.662}$ &
  $0.286^{\pm1.167}$ &
  $0.193^{\pm0.847}$ \\
\multicolumn{1}{|c|}{\multirow{-2}{*}{\textbf{ECL}}} &
  96 &
  $1.024^{\pm0.074}$ &
  $\textbf{0.022}^{\pm\textbf{0.02}}$ &
  $0.662^{\pm1.403}$ &
  $0.403^{\pm1.384}$ &
  $0.409^{\pm1.524}$ &
  {\color[HTML]{3465A4 } {$\underline{0.295}^{\underline{\pm1.456}}$}} \\ \hline
\multicolumn{1}{|c|}{} &
  3 &
  {\color[HTML]{3465A4 } {$\underline{0.051}^{\underline{\pm0.187}}$}} &
  $\textbf{0.031}^{\pm\textbf{0.151}}$ &
  $0.627^{\pm1.341}$ &
  $0.283^{\pm1.194}$ &
  $0.239^{\pm0.731}$ &
  $0.182^{\pm0.702}$ \\
\multicolumn{1}{|c|}{\multirow{-2}{*}{\textbf{ETTh1}}} &
  96 &
  $0.923^{\pm0.043}$ &
  $\textbf{0.018}^{\pm\textbf{0.008}}$ &
  $0.688^{\pm1.897}$ &
  $0.264^{\pm0.919}$ &
  $0.242^{\pm0.695}$ &
  {\color[HTML]{3465A4 } {$\underline{0.178}^{\underline{\pm0.749}}$}} \\ \hline
\multicolumn{1}{|c|}{} &
  3 &
  {\color[HTML]{3465A4 } {$\underline{0.036}^{\underline{\pm0.025}}$}} &
  $\textbf{0.02}^{\pm\textbf{0.017}}$ &
  $0.644^{\pm1.408}$ &
  $0.231^{\pm0.641}$ &
  $0.235^{\pm0.628}$ &
  $0.166^{\pm0.621}$ \\
\multicolumn{1}{|c|}{\multirow{-2}{*}{\textbf{Exchange}}} &
  96 &
  $0.938^{\pm0.065}$ &
  $\textbf{0.021}^{\pm\textbf{0.017}}$ &
  $0.622^{\pm1.209}$ &
  $0.238^{\pm0.608}$ &
  $0.241^{\pm0.649}$ &
  {\color[HTML]{3465A4 } {$\underline{0.169}^{\underline{\pm0.642}}$}} \\ \hline
\multicolumn{1}{|c|}{} &
  3 &
  {\color[HTML]{3465A4 } {$\underline{0.035}^{\underline{\pm0.024}}$}} &
  $\textbf{0.017}^{\pm\textbf{0.010}}$ &
  $0.631^{\pm1.214}$ &
  $0.235^{\pm0.703}$ &
  $0.233^{\pm0.666}$ &
  $0.176^{\pm0.706}$ \\
\multicolumn{1}{|c|}{\multirow{-2}{*}{\textbf{WTH}}} &
  96 &
  $0.927^{\pm0.049}$ &
  $\textbf{0.017}^{\pm\textbf{0.007}}$ &
  $0.661^{\pm1.66}$ &
  $0.244^{\pm0.697}$ &
  $0.249^{\pm0.701}$ &
  {\color[HTML]{3465A4 } {$\underline{0.176}^{\underline{\pm0.676}}$}} \\ \hline
\end{tabular}
\end{adjustbox}
\caption{\small Latency of Online learning models on Nvidia Jetson Nano($mean^{\pm std}$)}
\label{latency-gpu}
\end{center}
\end{table*}
\begin{table*}[h]
\begin{center}
\begin{adjustbox}{width=0.75\textwidth}
\begin{tabular}{cc|cccccc|}
\cline{3-8}
\multicolumn{2}{c|}{} &
  {\color[HTML]{5B277D} \textbf{AR-HDC}} &
  {\color[HTML]{5B277D} \textbf{Seq2Seq-HDC}} &
  {\color[HTML]{ACB20C} \textbf{FSNet}} &
  {\color[HTML]{ACB20C} \textbf{ER}} &
  {\color[HTML]{ACB20C} \textbf{DER++}} &
  {\color[HTML]{ACB20C} \textbf{OnlineTCN}} \\ \hline
\multicolumn{1}{|c|}{} &
  3 &
   {$3.58^{\pm0.257}$} &
  $\textbf{3.296}^{\pm\textbf{0.164}}$ &
  $4.021^{\pm0.465}$ &
  $3.362^{\pm0.121}$ &
  {\color[HTML]{3465A4}$\underline{3.344}^{\underline{\pm0.097}}$} &
  $3.363^{\pm0.264}$ \\
\multicolumn{1}{|c|}{\multirow{-2}{*}{\textbf{ECL}}} &
  96 &
  $5.501^{\pm0.501}$ &
  $\textbf{3.755}^{\pm\textbf{0.429}}$ &
  {\color[HTML]{3465A4 } {$\underline{4.141}^{\underline{\pm0.397}}$}} &
  $5.222^{\pm0.526}$ &
  $5.17^{\pm0.519}$ &
  $5.047^{\pm0.378}$ \\ \hline
\multicolumn{1}{|c|}{} &
  3 &
  $\textbf{3.184}^{\pm\textbf{0.109}}$ &
  {\color[HTML]{3465A4 } {$\underline{3.3}^{\underline{\pm0.017}}$}} &
  $4.188^{\pm0.303}$ &
  $3.318^{\pm0.129}$ &
  $3.234^{\pm0.056}$ &
  $3.302^{\pm0.211}$ \\
\multicolumn{1}{|c|}{\multirow{-2}{*}{\textbf{ETTh1}}} &
  96 &
  {\color[HTML]{3465A4 } {$\underline{3.523}^{\underline{\pm0.086}}$}} &
  $\textbf{3.393}^{\pm\textbf{0.287}}$ &
  $4.234^{\pm0.316}$ &
  $4.307^{\pm0.621}$ &
  $4.688^{\pm0.524}$ &
  $3.636^{\pm0.179}$ \\ \hline
\multicolumn{1}{|c|}{} &
  3 &
  $\textbf{3.395}^{\pm\textbf{0.211}}$ &
  $3.532^{\pm0.142}$ &
  $4.04^{\pm0.3}$ &
  $3.417^{\pm0.126}$ &
  $3.487^{\pm0.29}$ &
  {\color[HTML]{3465A4 } {$\underline{3.262}^{\underline{\pm0.134}}$}} \\
\multicolumn{1}{|c|}{\multirow{-2}{*}{\textbf{Exchange}}} &
  96 &
  $\textbf{3.539}^{\pm\textbf{0.089}}$ &
  {\color[HTML]{3465A4 } {$\underline{3.706}^{\underline{\pm0.318}}$}} &
  $4.289^{\pm0.351}$ &
  $4.474^{\pm0.596}$ &
  $4.337^{\pm0.449}$ &
  $3.73^{\pm0.043}$ \\ \hline
\multicolumn{1}{|c|}{} &
  3 &
  {\color[HTML]{3465A4 } {$\underline{3.157}^{\underline{\pm0.02}}$}} &
  $\textbf{3.118}^{\pm\textbf{0.125}}$ &
  $4.148^{\pm0.309}$ &
  $3.298^{\pm0.06}$ &
  $3.362^{\pm0.155}$ &
  $3.288^{\pm0.114}$ \\
\multicolumn{1}{|c|}{\multirow{-2}{*}{\textbf{WTH}}} &
  96 &
  $3.536^{\pm0.088}$ &
  {\color[HTML]{3465A4 } {$\underline{3.502}^{\underline{\pm0.406}}$}} &
  $4.265^{\pm0.291}$ &
  $4.394^{\pm0.545}$ &
  $4.243^{\pm0.423}$ &
  $\textbf{3.451}^{\pm\textbf{0.099}}$ \\ \hline
\end{tabular}
\end{adjustbox}
\caption{\small Power consumption of Online learning models on Nvidia Jetson Nano ($mean^{\pm std}$)}
\label{power-gpu}
\end{center}
\end{table*}

\section{Hardware Setup}
To evaluate the precision of TSF-HD and compare it against the baselines (as in Section \ref{precision} and Appendix \ref{power_results}), we trained our models (AR-HD \& Seq2Seq-HD) on an Nvidia RTX 3050 with 4GB of RAM. The baseline models except for GBRT \cite{friedman2001greedy} were trained on an Nvidia RTX A2000 with 12GB of RAM. GBRT \cite{friedman2001greedy} was trained on a CPU 11th Gen Intel(R) Core(TM) i7-11800H @ 2.30GHz. 

To evaluate the inference latency and power overhead of TSF-HD in comparison to the baselines, we conducted measurements on a RaspberryPi4 (Quad core Cortex-A72 (ARM v8) 64-bit SoC @ 1.8GHz) and on an Nvidia-Jetson Nano with 4 GB of RAM. The power usage of the RaspberryPi4 was measured with a USB Powermeter placed between the board and the Power Supply Unit (PSU). For the Nvidia-Jetson the power was measured internally using the python library \texttt{Jtop}. We note that some baselines such as Informer \cite{zhou2021informer} are unable to fit on a Raspberry Pi and therefore were not considered for inference latency and power measurement.

\section{Latency and Power Results}\label{appendices_hw}

Our complete findings for latency and power on a Raspberry Pi are detailed in Table \ref{latency-cpu} (latency) and Table \ref{power-cpu} (power) for four different datasets and two different values of $\tau$ representing short and long-term forecast horizons, comparing the two TSF-HD models with our online learning baselines. For short-term TSF ($\tau=3$), AR-HDC demonstrates the lowest latency, followed by Seq2Seq-HDC. Notably, both models maintain a nearly constant latency with minimal standard deviation.
For long-term TSF, AR-HDC falls short, exhibiting the highest latency and reduced power efficiency. This is attributed to its use of a loop for prediction and update phases, updating the model $\tau \gg 1$ times during the update loop. In contrast, Seq2Seq-HDC outperforms the others in both inference latency and power efficiency thanks to its use of one-shot prediction.
Seq2Seq-HDC is seen to be faster than baseline models and AR-HDC on edge GPUs for both short and long-term TSF. On edge CPUs, AR-HDC is faster and more power-efficient the other models, but only for short-term forecasting scenarios. 

Tables \ref{latency-gpu} and \ref{power-gpu} present similar results, comparing latency and power consumption of AR-HDC and Seq2Seq-HDC with the online learning baselines over four datasets and two values of $\tau$ for short- and long-term TSF on an Nvidia Jetson Nano edge GPU. Unlike the ARM CPU of the Raspberry Pi, the edge GPU is better suited for parallel computing, enabling faster matrix computations. This advantage is evident as Seq2Seq-HDC outpaces AR-HDC in short-term TSF as well as long-term TSF on this platform. From a power consumption standpoint, AR-HDC and Seq2Seq-HDC rank as the most and second-most efficient models, respectively, except in the Exchange dataset for short-term forecasting (where Online TCN is the second most power-efficient) and Weather dataset for long-term forecasting (where OnlineTCN is the most power-efficient).
To summarize, the Seq2Seq-HDC model proves more efficient than baseline models and AR-HDC on edge GPUs for both short and long-term TSF. Conversely, on edge CPUs, AR-HDC demonstrates lower overhead than Seq2Seq-HDC and other baseline models, but only in short-term forecasting scenarios. 

\section{Reproducibility Details}    
The experiments are performed using 5 different random seed values, which are $2019$, $2020$, $2021$, $2022$ and $2023$.
The learning rate used are summarized in the table \ref{learning_rate}. We note that the experiments can be replicated using the provided GitHub repository.
\begin{table*}[ht]
\begin{center}
\begin{adjustbox}{center, width=0.75\textwidth}
\begin{tabular}{c|cccccccc|}
\cline{2-9}
                                  & Exchange & ECL  & ETTh1 & ETTh2 & ETTm1 & ETTm2 & WTH  & ILI  \\ \hline
\multicolumn{1}{|c|}{Seq2Seq-HDC} & 1e-3     & 2e-5 & 1e-4  & 1e-4  & 1e-4  & 1e-4  & 1e-4 & 1e-3 \\
\multicolumn{1}{|c|}{AR-HDC}      & 1e-4     & 4e-5 & 5e-5  & 5e-5  & 5e-5  & 5e-5  & 4e-4 & 1e-3 \\ \hline
\end{tabular}
\end{adjustbox}
\end{center}
\caption{Learning rate parameter values}
\label{learning_rate}
\end{table*}


\end{document}